\documentclass[11pt]{article}

\usepackage{acl}
\usepackage{times}
\usepackage{latexsym}
\usepackage[T1]{fontenc}
\usepackage[utf8]{inputenc}
\usepackage{microtype}
\usepackage{inconsolata}
\usepackage{graphicx}
\usepackage{amsmath}
\usepackage{amssymb}
\usepackage{booktabs}
\usepackage{multirow}
\usepackage{array}
\usepackage{makecell}
\usepackage{xcolor}

\usepackage[most]{tcolorbox}
\tcbuselibrary{listings,breakable,skins}
\usepackage{xcolor}

\definecolor{boxbg}{HTML}{F8FAFC}
\definecolor{boxframe}{HTML}{CBD5E1}
\definecolor{boxtitle}{HTML}{E2E8F0}
\newcolumntype{P}[1]{>{\raggedright\arraybackslash}p{#1}}

\newtcblisting{promptbox}[2][]{
  enhanced,
  listing only,
  width=\linewidth,
  colback=boxbg,
  colframe=boxframe,
  colbacktitle=boxtitle,
  coltitle=black,
  title={#2},
  fonttitle=\bfseries\small,
  boxrule=0.45pt,
  arc=1.2mm,
  left=1.8mm,
  right=1.8mm,
  top=1.2mm,
  bottom=1.2mm,
  before skip=1.5mm,
  after skip=1.5mm,
  listing options={
    basicstyle=\ttfamily\scriptsize,
    breaklines=true,
    breakatwhitespace=false,
    columns=fullflexible,
    keepspaces=true,
    showstringspaces=false,
    tabsize=2
  },
  #1
}

\newtcolorbox{papertextbox}[2][]{
  enhanced,
  breakable,
  width=\linewidth,
  colback=boxbg,
  colframe=boxframe,
  colbacktitle=boxtitle,
  coltitle=black,
  title={#2},
  fonttitle=\bfseries\small,
  fontupper=\scriptsize,
  boxrule=0.45pt,
  arc=1.2mm,
  left=1.8mm,
  right=1.8mm,
  top=1.2mm,
  bottom=1.2mm,
  before skip=1.5mm,
  after skip=1.5mm,
  #1
}

\title{LivingRAG: Augmenting Graph RAG with Experience}
\author{
  \textbf{Yuzhuo Cui\textsuperscript{1,2}},
  \textbf{Zongye Zhang\textsuperscript{1,2}},
  \textbf{Qingjie Liu\textsuperscript{1,2,*}} \\
  \textsuperscript{1}State Key Laboratory of Virtual Reality Technology and Systems, \\
  Beihang University, Beijing, China \\
  \textsuperscript{2}Hangzhou Innovation Institute, Beihang University, \\
  Hangzhou, Zhejiang, China \\
  \texttt{\{cyz020403,zhangzongye,qingjie.liu\}@buaa.edu.cn} \\
  \small{\textsuperscript{*}Corresponding author}}

\begin{document}
\maketitle

\begin{abstract}
Graph-based RAG improves multi-hop question answering by organizing evidence as a knowledge graph. 
However, most existing RAG systems process each query in isolation and discard useful reasoning from the LLM's response after inference.
As a result, later related queries need to retrieve evidence and reason from scratch.
We propose LivingRAG, a Graph RAG framework with writable and reusable reasoning experience. 
LivingRAG adds a writable experience store to a graph-based retrieval backbone, enabling verified experiences to be reused during inference in two ways.
Stored graph signals help retrieval find entities and passages that were useful in earlier related queries.
Stored summaries provide a reference reasoning pattern for answer generation.
We analyze online QA streams and find reusable signals from shared entities, graph neighborhoods, and question templates. 
Experiments on multi-hop QA benchmarks show that LivingRAG improves accuracy over strong RAG baselines and reduces completion-token use when relevant prior experience is reused.
\end{abstract}

\section{Introduction}
Large language models perform well on many language understanding and reasoning tasks.
However, they still face limits in knowledge-intensive tasks.
They may hallucinate facts.
They may also lack access to domain-specific or up-to-date knowledge.
Retrieval-augmented generation addresses these limits by grounding model outputs in external documents~\citep{lewis2020retrieval,karpukhin2020dense,gao2023retrieval}.

Graph-based RAG further improves retrieval for complex questions.
It represents passages, entities, and their relations in a graph.
The graph structure helps the retriever find multi-hop evidence that dense retrieval may miss.
Recent systems show that graph structure is useful for global context and entity-level reasoning~\citep{graphRAG,lightRAG,hippoRAG2,linearRAG,gfmRAG}.
However, online QA streams raise a further question.
When related queries arrive over time, should the system reuse useful reasoning from earlier queries?

Despite this progress, most RAG and Graph RAG systems still answer each query independently at inference time~\citep{graphRAG,lightRAG,hippoRAG2,linearRAG,gfmRAG}.
These methods retrieve evidence for the current query and then generate an answer, but they discard most query-level reasoning signals after inference.
As a result, later related queries often need to retrieve and reason again from scratch.
This read-only inference process limits reuse in online QA streams.

This limitation matters because useful reuse is not limited to repeated entities.
Two questions may visit nearby regions of the retrieval graph.
They may also follow the same reasoning pattern with different entities.
For example, two comparison questions may both ask which person was born earlier, even when the people and evidence passages differ.
Therefore, a useful online Graph RAG system should preserve more than final answers.
It should preserve graph-neighborhood priors and reusable reasoning patterns.

To address this problem, we introduce LivingRAG, a Graph RAG framework with writable and reusable reasoning experience.
LivingRAG adds a writable experience store to a graph-based retrieval backbone.
After each query, it can write a compact experience if the experience is grounded and novel.
Each stored experience keeps the source query, activated graph entities, a short reasoning summary, and validation metadata.
During later inference, LivingRAG reuses stored experiences in two ways.
Activation maps guide graph retrieval toward useful evidence neighborhoods.
Reasoning summaries provide scaffolds for answer generation.

A writable experience store also introduces a risk.
If unsupported reasoning is written into the store, later queries may reuse it and amplify the error.
LivingRAG reduces this risk with quality gates.
A candidate is stored only when its claims are supported by retrieved evidence and when it is not redundant with stored experiences.
Thus, LivingRAG can return an answer without storing low-quality experience.

We evaluate LivingRAG on multi-hop QA benchmarks.
The experience store is initialized at the beginning of each run and updated online during that run.
Across these benchmarks, LivingRAG improves accuracy over strong RAG baselines.
It also reduces completion-token use and estimated API cost when later queries retrieve useful prior experience.

Our contributions are:
(1) We introduce LivingRAG, a Graph RAG framework with writable and reusable reasoning experience. It stores grounded and novel experiences, and reuses them through activation maps for retrieval and reasoning summaries for generation.
(2) We analyze online QA streams and find reusable signals from shared entities, graph neighborhoods, and question templates. We show that LivingRAG can reuse these signals in both multi-hop QA and support-style QA settings.
(3) We show that LivingRAG improves accuracy over strong RAG baselines. It also reduces completion-token use and estimated API cost when relevant prior experience is reused.

\section{Related Work}

Retrieval-augmented generation grounds language models in external documents and reduces reliance on parametric knowledge~\citep{lewis2020retrieval,karpukhin2020dense,gao2023retrieval}.
Early systems usually retrieve independent text chunks.
Recent Graph RAG methods add graph structure to support complex retrieval.
GraphRAG builds entity graphs and community summaries~\citep{graphRAG}.
LightRAG combines graph indexing with vector retrieval~\citep{lightRAG}.
HippoRAG and HippoRAG2 use graph-based associative retrieval~\citep{hippoRAG,hippoRAG2}.
LinearRAG uses a lightweight passage-sentence-entity graph for efficient retrieval~\citep{linearRAG}.
These systems show that graph structure helps multi-hop evidence search.
LivingRAG builds on this direction, but studies a different problem.
It asks how a Graph RAG system can write and reuse verified reasoning experience across an online query stream.

Several methods also improve RAG with reasoning or memory.
Reasoning-enhanced RAG methods use decomposition, intermediate notes, self-reflection, or tool use to improve single-query reasoning~\citep{trivedi2023interleaving,jiang2023active,asai2024selfRAG,chainofNote,logicRAG}.
Memory-based agents store past observations or feedback for later decisions~\citep{reflexion,selfRefine}.
Recent memory-augmented RAG systems further show that reusable memory can support retrieval.
REMem builds a hybrid episodic memory graph for language-agent histories and uses an agentic retriever for episodic recollection and reasoning~\citep{shu2026remem}.
MemoRAG builds a global memory for long-context processing and uses it to produce clues for retrieval~\citep{MemoRAG}.
LivingRAG is complementary to these works.
It focuses on online Graph RAG for QA streams.
Its written unit is a verified reasoning experience, not a raw interaction history or a global long-context memory.
Each experience stores graph activation information and a compact reasoning scaffold.
It is written only after grounding and novelty checks.
The experience is not only retrieved as text.
It also changes the initial graph activation used by the retriever.

\section{Method}
\label{sec:method}

\subsection{Motivating Analysis}
\label{sec:motivating-analysis}

LivingRAG is designed for online QA streams.
In this setting, the system answers queries one by one.
For a query $q_t$, only earlier queries $q_1,\ldots,q_{t-1}$ can be used as reuse sources.
This causal setup matches an online experience store and avoids using future information.

We measure three reuse signals.
Entity reuse occurs when the current query shares at least one seed entity with an earlier query.
Graph-neighborhood reuse occurs when the current query and an earlier query visit nearby regions of the retrieval graph.
We measure this signal with a fixed LinearRAG retriever.
Two queries are counted as graph-neighborhood reuse when their retrieved passages or filtered context entities have Jaccard similarity of at least 0.05.
Template reuse occurs when two queries have no seed-entity overlap but have similar masked questions.
We mask entity mentions, quoted spans, and numeric as Section~\ref{sec:experience-augmented-inference}.
We then compute TF-IDF cosine similarity between the masked questions.
Two queries are counted as template reuse when the score is at least 0.35.

\begin{table}[t]
\centering
\small
\setlength{\tabcolsep}{3pt}
\caption{Reuse signals in online QA streams.
Only earlier queries are used as reuse sources.
All values are percentages.}
\label{tab:reuse_opportunities}
\begin{tabular*}{\columnwidth}{@{\extracolsep{\fill}}lccc@{}}
\toprule
\textbf{Dataset} & \shortstack{\textbf{Entity}\\\textbf{reuse}} & \shortstack{\textbf{Graph-neigh.}\\\textbf{reuse}} & \shortstack{\textbf{Template}\\\textbf{reuse}} \\
\midrule
2Wiki          & 4.30  & 99.70 & 96.30 \\
HotpotQA       & 26.53 & 99.10 & 21.82 \\
MuSiQue        & 40.34 & 95.60 & 39.14 \\
MuSiQue-full   & 54.10 & 96.32 & 54.30 \\
WixQA          & 76.69 & 80.45 & 16.04 \\
\bottomrule
\end{tabular*}
\end{table}

Table~\ref{tab:reuse_opportunities} shows that reusable signals exist across online QA streams, but they appear in different forms.
Direct entity reuse is limited in most datasets, so an entity-only cache would miss many opportunities.
In contrast, graph-neighborhood reuse is common across datasets.
This suggests that later queries often need evidence from regions that are close to earlier retrieval paths.

Template reuse is also useful, but it is more dataset-dependent.
It is stronger in formulaic multi-hop streams and weaker in more diverse support-style questions.
These results motivate LivingRAG to reuse experience through two paths.

\subsection{Overview}
\label{sec:method-overview}

\begin{figure*}[t]
\centering
\includegraphics[width=\textwidth]{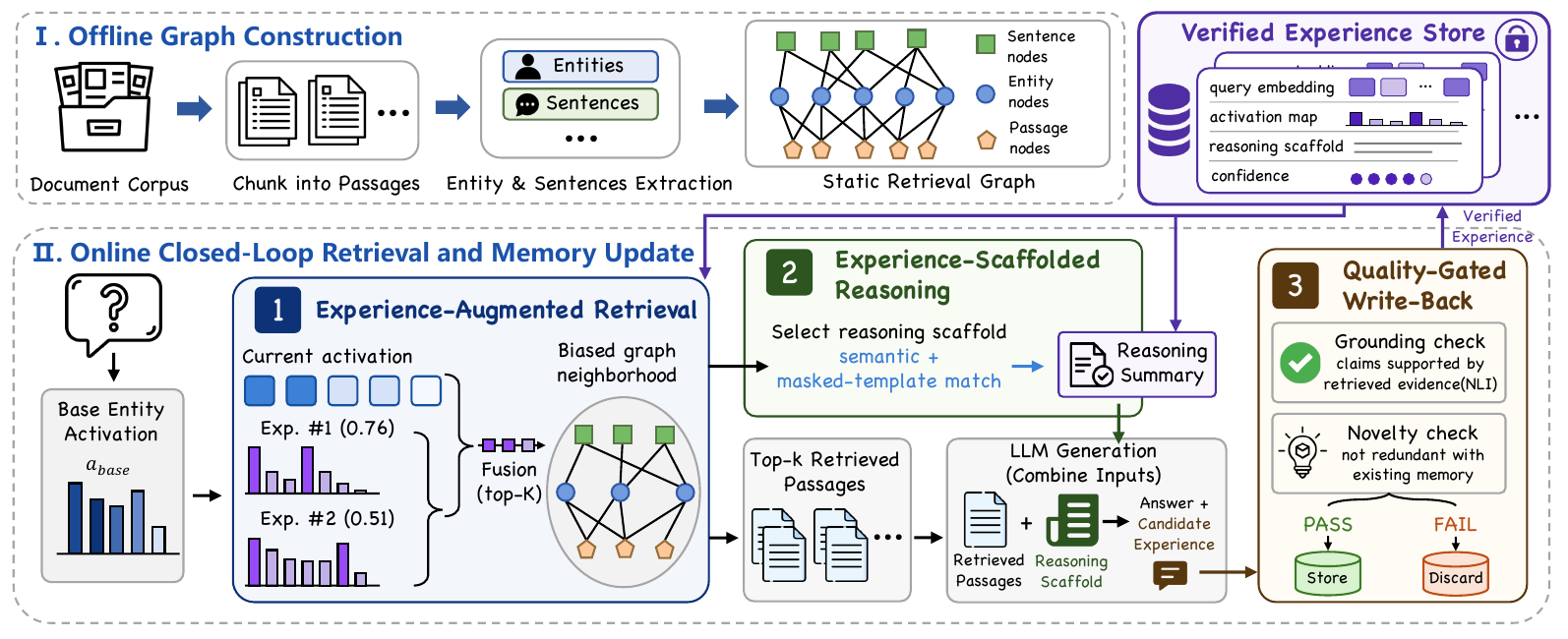}
\caption{Overview of LivingRAG.
The framework first constructs a static retrieval graph offline.
It then performs experience-augmented retrieval, scaffolded generation, and write-back through quality gates.}
\label{fig:livingrag_overview}
\end{figure*}

LivingRAG extends a graph-based RAG system with a writable experience store.
The base retriever follows LinearRAG~\cite{linearRAG}.
Given a document corpus $\mathcal{D}$, it builds a retrieval graph with passage nodes, sentence nodes, and entity nodes.
The graph is constructed with lightweight entity extraction and does not require LLM calls during indexing.
For each query, LinearRAG first activates query-related entities, propagates relevance over the sentence-entity graph, and then ranks passages with Personalized PageRank over the passage-entity graph.
LivingRAG keeps this retrieval backbone unchanged.
It adds a writable experience store during an online QA stream.
LivingRAG does not reuse previous answers as a response cache.
It reuses verified activation maps and reasoning scaffolds during retrieval and generation.
Figure~\ref{fig:livingrag_overview} gives an overview of the full workflow.

At time step $t$, the system receives a query $q_t$ and has access to a static retrieval graph $\mathcal{G}$ and an experience store $\mathcal{E}_t$.
A standard graph-based RAG system answers each query independently:
\begin{equation}
a_t = \mathrm{LLM}\bigl(q_t, \mathcal{R}(q_t, \mathcal{G})\bigr),
\end{equation}
where $\mathcal{R}$ retrieves passages from the graph.
LivingRAG instead uses stored experiences during both retrieval and generation:
\begin{equation}
\begin{aligned}
a_t = \mathrm{LLM}\bigl(&q_t,
\mathcal{R}_{\mathrm{exp}}(q_t, \mathcal{G}, \mathcal{E}_t),\\
&\mathcal{S}_{\mathrm{exp}}(q_t, \mathcal{E}_t)\bigr).
\end{aligned}
\end{equation}
Here $\mathcal{R}_{\mathrm{exp}}$ is experience-augmented graph retrieval, and $\mathcal{S}_{\mathrm{exp}}$ is an optional reasoning scaffold selected from the experience store.
If no useful stored experience is found, LivingRAG reduces to the base graph RAG system.

After answering query $q_t$, LivingRAG forms a \emph{candidate experience} $C_t$.
The candidate contains the query, retrieved passages, generated explanation, and final answer.
LivingRAG then decides whether the candidate should be written to the store.
If the candidate passes the grounding and novelty checks, it becomes a \emph{verified experience} $e_t$.
We use $C_t$ for the candidate before validation and $e_t$ for the verified experience after validation.
The experience-store update is:
\begin{equation}
\mathcal{E}_{t+1} =
\begin{cases}
\mathcal{E}_t \cup \{e_t\}, &
\text{if } \mathrm{Store}(C_t, \mathcal{E}_t)=1, \\
\mathcal{E}_t, &
\text{otherwise}.
\end{cases}
\end{equation}
The function $\mathrm{Store}$ checks whether the candidate is grounded in retrieved evidence and sufficiently different from stored experiences.
This write-back mechanism lets LivingRAG accumulate reusable experiences without changing model parameters.

\subsection{Experience Store}

The basic unit of the store is a verified experience.
After this definition, we use experience as a short form when the meaning is clear.
Each verified experience is a compact form of one candidate experience that passed validation:
\begin{equation}
e_t =
\bigl(
q_t,
\mathbf{v}_t,
\tilde{\mathbf{a}}_t,
\mathrm{sum}_t,
a_t,
ts_t,
\mathrm{conf}_t
\bigr).
\end{equation}
Here $q_t$ is the source query, $\mathbf{v}_t=\mathrm{Emb}(q_t)$ is its query embedding, 
$\tilde{\mathbf{a}}_t$ is a sparse activation map over graph entities, $\mathrm{sum}_t$ is a compact reasoning summary, 
$a_t$ is the final answer, $ts_t$ is the timestamp, and $\mathrm{conf}_t$ is the grounding confidence.

The activation map is the main graph-space representation of a verified experience.
After the base retriever finishes entity propagation for query $q_t$, it produces a final entity activation vector $\mathbf{a}^{\mathrm{final}}_{q_t}$.
LivingRAG sparsifies this vector:
\begin{equation}
\tilde{\mathbf{a}}_t =
\mathrm{sparsify}(\mathbf{a}^{\mathrm{final}}_{q_t}, \epsilon),
\end{equation}
where entries below threshold $\epsilon$ are set to zero.
The sparse activation map stores which graph entities were useful for the candidate and how strongly they were activated.
Unlike a plain entity list, the map preserves graded graph-neighborhood information and can be reused directly in later graph retrieval.

The compact reasoning summary becomes a reasoning scaffold during later generation.
It is not the full raw reasoning trace.
To keep the prompt short, LivingRAG selects a small number of key sentences from the generated explanation.
The selection is anchored by the most activated entities in $\mathbf{a}^{\mathrm{final}}_{q_t}$.
Sentences that mention more activated entities are ranked higher, and the selected sentences are kept in their original order.
The stored summary is:
\begin{equation}
\mathrm{sum}_t =
\bigl(q_t, \mathrm{KeySent}(r_t, \mathbf{a}^{\mathrm{final}}_{q_t}), a_t\bigr),
\end{equation}
where $r_t$ is the generated explanation.
The compact summary preserves the main reasoning pattern while adding few tokens to future prompts.

\subsection{Experience-Augmented Retrieval and Scaffolded Generation}
\label{sec:experience-augmented-inference}

LivingRAG uses the experience store in two places.
First, it uses activation maps to guide graph retrieval.
Second, it uses compact summaries as reasoning scaffolds for answer generation.
The two paths are related but separate.
The retrieval path transfers graph-neighborhood information.
The generation path transfers a reusable reasoning pattern.

For graph retrieval, LivingRAG first matches the current query with stored experiences.
Each stored experience is scored by combining semantic similarity and activation-map similarity:
\begin{equation}
\begin{aligned}
s_k^{\mathrm{ret}}
= {}&
\alpha
\cos\bigl(\mathrm{Emb}(q_t), \mathbf{v}_k\bigr)\\
&+
(1-\alpha)
\cos\bigl(\mathbf{a}^{\mathrm{base}}_{q_t}, \tilde{\mathbf{a}}_k\bigr).
\end{aligned}
\end{equation}
Here $\mathbf{a}^{\mathrm{base}}_{q_t}$ is the initial entity activation vector produced from the current query before stored experiences are used.
The first term captures surface-level query similarity.
The second term captures whether the current query starts near an evidence neighborhood that was useful before.
LivingRAG retrieves the top-$K$ experiences according to $s_k^{\mathrm{ret}}$.

The retrieved activation maps are then fused into the initial entity activation vector:
\begin{equation}
\mathbf{a}^{0}_{q_t}
=
\mathrm{normalize}
\left(
\mathbf{a}^{\mathrm{base}}_{q_t}
+
\beta
\sum_{e_k \in \mathcal{N}^{\mathrm{ret}}_t}
w_k \tilde{\mathbf{a}}_k
\right).
\end{equation}
We first compute a softmax weight over the selected experiences:
\begin{equation}
\bar{w}_k =
\frac{\exp(s_k^{\mathrm{ret}}/\tau_s)}
{\sum_{j \in \mathcal{N}^{\mathrm{ret}}_t}
\exp(s_j^{\mathrm{ret}}/\tau_s)}.
\end{equation}
We then scale this weight by the confidence of the stored experience:
\begin{equation}
w_k = \bar{w}_k \cdot \mathrm{conf}_k.
\end{equation}
The scaled weights are not renormalized.
Thus, low-confidence experiences have less influence on the fused activation vector.
The resulting vector $\mathbf{a}^{0}_{q_t}$ replaces the original initial activation vector of the base retriever.
All later retrieval steps are unchanged.
When the experience store is empty, or when the current query has no seed entities for graph fusion, LivingRAG falls back to the base retriever.

For generation, LivingRAG may also select a reasoning scaffold.
Scaffold selection can use a template signal because the selected scaffold is added only to the LLM prompt and does not inject historical entities into graph retrieval.
For a query $q$, we define a masked template:
\begin{equation}
T(q) = \mathrm{Mask}(q),
\end{equation}
where entity mentions, quoted spans, and numeric expressions are replaced with placeholders.
The scaffold score is:
\begin{equation}
\begin{aligned}
s_k^{\mathrm{scaf}}
= {}&
(1-\lambda)s_k^{\mathrm{ret}}\\
&+
\lambda
\cos\bigl(\mathrm{Emb}(T(q_t)), \mathrm{Emb}(T(q_k))\bigr).
\end{aligned}
\end{equation}
When stored experiences are available, LivingRAG selects the highest-scoring experience as the scaffold:
\begin{equation}
e_t^{\mathrm{scaf}}
=
\operatorname*{arg\,max}_{e_k \in \mathcal{E}_t}
s_k^{\mathrm{scaf}}.
\end{equation}
If the experience store is empty, or if no useful scaffold is selected, no scaffold is inserted.
In the special case $\lambda=0$, scaffold selection reduces to the highest-scoring experience from experience-augmented retrieval.

The LLM receives the retrieved passages and, when available, the selected scaffold:
\begin{equation}
\mathrm{Context}_t
=
\{p_1,\ldots,p_k\}
\cup
\{\mathrm{sum}(e_t^{\mathrm{scaf}})\}.
\end{equation}
The scaffold shows how a related problem was solved.
The model can still use the current passages to produce a different answer when the new query requires different evidence.

\subsection{Experience Write-Back}
\label{sec:experience-write-back}

After generation, LivingRAG checks whether the new candidate experience should be stored.
LivingRAG separates write-back validation from answer generation.
The system can return an answer even when the candidate is not stored.
This separation prevents unsupported or redundant reasoning from entering the experience store.

The first criterion is factual grounding.
LivingRAG extracts atomic claims from the available reasoning trace and the final answer.
Each claim is checked against the retrieved passages with an NLI model:
\begin{equation}
g_i =
\max_{p_j \in \mathcal{D}^{\mathrm{used}}_t}
P_{\mathrm{NLI}}(\mathrm{entail} \mid p_j, cl_i).
\end{equation}
The grounding score is:
\begin{equation}
\mathrm{GS}(C_t)
=
\frac{1}{M}
\sum_{i=1}^{M}
\mathbb{1}[g_i \geq \tau_{\mathrm{nli}}].
\end{equation}
If $\mathrm{GS}(C_t) < \tau_{\mathrm{gs}}$, the candidate is not stored.
The grounding score measures evidential support.
It does not explicitly measure whether a candidate will be useful for future questions.

The second criterion is novelty.
LivingRAG avoids storing candidates that are near duplicates of stored experiences.
Novelty is measured before experience fusion, using the current query embedding and its base activation vector:
\begin{equation}
\begin{aligned}
\eta_k =
\alpha
\cos\bigl(\mathrm{Emb}(q_t), \mathbf{v}_k\bigr)\\
{}+
(1-\alpha)
\cos\bigl(\mathbf{a}^{\mathrm{base}}_{q_t}, \tilde{\mathbf{a}}_k\bigr),
\end{aligned}
\end{equation}
\begin{equation}
\mathrm{NS}(C_t,\mathcal{E}_t)
=
1 - \max_{e_k \in \mathcal{E}_t} \eta_k.
\end{equation}
Using the base activation vector avoids circularity.
A new candidate should not look redundant only because a previous stored experience influenced retrieval.
If $\mathrm{NS}(C_t,\mathcal{E}_t) < \tau_{\mathrm{novel}}$, LivingRAG returns the answer but does not store the candidate.

The write-back rule is:
\begin{equation}
\begin{aligned}
\mathrm{Store}(C_t,\mathcal{E}_t)=1
\quad \text{iff} \quad
\mathrm{GS}(C_t) \geq \tau_{\mathrm{gs}}\\
{}\land\
\mathrm{NS}(C_t,\mathcal{E}_t) \geq \tau_{\mathrm{novel}}.
\end{aligned}
\end{equation}
The rule defines the storage condition.
In implementation, LivingRAG checks novelty before grounding to avoid unnecessary NLI calls.
Appendix~\ref{app:eqs-gates} reports gate dynamics and examples of accepted and rejected candidates.
For accepted candidates, the confidence score is set to the grounding score $\mathrm{conf}_t = \mathrm{GS}(C_t)$.
The new verified experience is then appended to the store and can be used by later queries.

\begin{table*}[t]
\centering
\scriptsize
\setlength{\tabcolsep}{2pt}
\caption{Overall accuracy (\%) on QA benchmarks.
Cont.-Acc. denotes Contain-Match Accuracy.
WixQA is reported only for LinearRAG and LivingRAG; unavailable entries are marked with --.
The best result is marked in \textbf{bold}, and the second-best result is underlined.}
\label{tab:overall_accuracy}
\resizebox{\textwidth}{!}{%
\begin{tabular}{lccccccccc}
\toprule
\multirow{2}{*}{\textbf{Method}}
& \multicolumn{2}{c}{\textbf{2Wiki}}
& \multicolumn{2}{c}{\textbf{HotpotQA}}
& \multicolumn{2}{c}{\textbf{MuSiQue}}
& \multicolumn{2}{c}{\textbf{MuSiQue-full}}
& \textbf{WixQA} \\
\cmidrule(lr){2-3} \cmidrule(lr){4-5} \cmidrule(lr){6-7} \cmidrule(lr){8-9} \cmidrule(lr){10-10}
& Cont.-Acc. & LLM-Acc.
& Cont.-Acc. & LLM-Acc.
& Cont.-Acc. & LLM-Acc.
& Cont.-Acc. & LLM-Acc.
& LLM-Acc. \\
\midrule
Vanilla RAG (Top-5)~\citep{karpukhin2020dense} & 58.80 & 59.40 & 65.20 & 77.10 & 34.00 & 43.70 & 27.35 & 41.33 & -- \\
LightRAG~\citep{lightRAG}                      & 60.60 & 48.20 & 62.20 & 71.70 & 39.20 & 49.10 & 29.83 & 44.89 & -- \\
GFM-RAG~\citep{gfmRAG}                          & 75.50 & 77.80 & 72.50 & 86.10 & 42.70 & 52.50 & 37.28 & 47.75 & -- \\
HippoRAG2~\citep{hippoRAG2}                     & 76.50 & 68.00 & 71.20 & 84.60 & 45.40 & 52.90 & 40.92 & 44.06 & -- \\
LinearRAG~\citep{linearRAG}                     & \underline{79.60} & \underline{84.40} & \underline{72.90} & \underline{86.80} & \underline{48.50} & \underline{60.70} & \underline{41.00} & \underline{52.71} & \underline{66.00} \\
\midrule
\textbf{LivingRAG (Ours)} & \textbf{82.70} & \textbf{87.00} & \textbf{73.20} & \textbf{88.00} & \textbf{50.80} & \textbf{64.40} & \textbf{44.39} & \textbf{58.42} & \textbf{70.25} \\
\bottomrule
\end{tabular}
}
\end{table*}

\subsection{Complexity}
\label{sec:complexity}

LivingRAG preserves the static graph construction and graph retrieval backbone of LinearRAG.
Therefore, the corpus-side complexity remains unchanged.
For each query, the added retrieval cost comes from matching the query against stored experiences.
If the experience store contains $|\mathcal{E}_t|$ experiences and the average number of nonzero entries in an activation map is $\bar{z}$, the added retrieval cost is $O(|\mathcal{E}_t|d + |\mathcal{E}_t|\bar{z})$,
where $d$ is the embedding dimension.
Since only the top-$K$ experiences are fused, and $K$ is fixed, the experience-augmented retrieval overhead is linear in the store size and independent of the corpus graph size.

Write-back validation is also selective.
Novelty scoring uses the same lightweight store scan.
Grounding verification is executed only for candidates that pass the novelty gate.
If the candidate contains $M$ claims and the validator keeps $K_g$ candidate evidence sentences per claim, the write-back overhead is:
\begin{equation}
\begin{aligned}
T_{\mathrm{write}}(q_t)
=
{}&
O(|\mathcal{E}_t|(d+\bar{z}))\\
&+
\mathbb{1}_{\mathrm{novel}}\,O(MK_g).
\end{aligned}
\end{equation}
Thus, the NLI grounding cost is paid only for candidates that pass novelty filtering.
The NLI cost is used for write-back validation.
It does not change the base graph index and does not require additional LLM calls.

The additional storage is compact.
Each stored experience contains a dense embedding, a sparse activation map, a short reasoning scaffold, and metadata.
The experience-store overhead is $O(|\mathcal{E}_t|(d+\bar{z}+\bar{\ell}))$, where $\bar{\ell}$ is the average scaffold length.
Appendix~\ref{app:complexity-analysis} provides the full time and space analysis.

When useful stored experiences are retrieved, LivingRAG can reduce generation cost in two ways.
Activation maps can move retrieval toward relevant evidence earlier, and scaffolds can reduce the reasoning the LLM must reconstruct from scratch.
We write the expected token cost as:
\begin{equation}
\begin{aligned}
\mathbb{E}[\mathrm{Cost}(t)]
= {}&
(1-f_{\mathrm{exp}}(t))C_{\mathrm{base}}
\\
&+
f_{\mathrm{exp}}(t)C_{\mathrm{exp}},
\end{aligned}
\end{equation}
where $f_{\mathrm{exp}}(t)$ is the fraction of queries with useful support from stored experiences.
Since $C_{\mathrm{exp}}$ depends on the quality of retrieved experiences, LivingRAG does not assume that cost always decreases.
It predicts lower cost when later queries can reuse relevant stored experiences, which we test in the experiments.

\section{Experiments}
\label{sec:experiments}

\subsection{Overall Accuracy on QA Benchmarks}

We first evaluate whether LivingRAG improves end-to-end QA accuracy over representative RAG and GraphRAG baselines.
Later sections isolate experience reuse, token use, and online efficiency.

\paragraph{Setup.}
We evaluate all methods on four multi-hop QA streams: 2WikiMultiHopQA, HotpotQA, MuSiQue, and MuSiQue-full.
MuSiQue-full uses the 2,417-question answerable stream described in Appendix~\ref{app:dataset-details}.
We additionally report WixQA only for LinearRAG and LivingRAG, since its LLM-based entity extraction is not suitable for the other baselines.
Appendix~\ref{app:wixqa-processing} describes these choices.
We compare LivingRAG with Vanilla RAG~\citep{karpukhin2020dense}, LightRAG~\citep{lightRAG}, GFM-RAG~\citep{gfmRAG}, HippoRAG2~\citep{hippoRAG2}, and LinearRAG~\citep{linearRAG}.
All methods use Qwen3.6 Plus for answer generation.
We report Contain-Match Accuracy and LLM-Evaluation Accuracy; WixQA uses only LLM-Evaluation Accuracy.

For LivingRAG, the experience store is initialized as empty at the beginning of each dataset and updated online during evaluation.
Only earlier queries can contribute experience to later queries, following the online setting defined in Section~\ref{sec:motivating-analysis}.

Table~\ref{tab:overall_accuracy} shows that LivingRAG achieves the best overall accuracy on the four multi-hop benchmarks and in the WixQA comparison against LinearRAG.
Under our evaluation setting, the writable experience store does not weaken the base QA ability of the graph-RAG backbone.
Since LivingRAG builds retrieval framework based on LinearRAG, later experiments use LinearRAG as the main baseline to better isolate the effect of experience reuse.
Appendix~\ref{app:ppr-transfer} reports a retrieval-level transfer test with HippoRAG2's entity-passage PPR.

\providecommand{\tokinc}[1]{\textcolor{red}{#1}}
\providecommand{\tokdec}[1]{\textcolor{green!50!black}{#1}}
\begin{table*}[t]
\centering
\small
\setlength{\tabcolsep}{3pt}
\caption{Token consumption and cost of LinearRAG and LivingRAG.
Prompt and completion tokens are per-query averages. 
Rel. $\Delta$ columns report $(\mathrm{LivingRAG}-\mathrm{LinearRAG})/\mathrm{LinearRAG}$. 
Negative values indicate reductions. Run cost is computed with Qwen3.6 Plus pricing: \$0.50 per million prompt tokens and \$3.00 per million completion tokens.}
\label{tab:token_efficiency}
\begin{tabular}{lrrrrrrrrrr}
\toprule
\multirow{2}{*}{\textbf{Dataset}} & \multirow{2}{*}{\textbf{\# Queries}}
& \multicolumn{3}{c}{\textbf{Prompt tokens}}
& \multicolumn{3}{c}{\textbf{Completion tokens}}
& \multicolumn{3}{c}{\textbf{Run cost (USD)}} \\
\cmidrule(lr){3-5} \cmidrule(lr){6-8} \cmidrule(lr){9-11}
& & \textbf{Linear} & \textbf{Living} & \textbf{Rel. $\Delta$}
& \textbf{Linear} & \textbf{Living} & \textbf{Rel. $\Delta$}
& \textbf{Linear} & \textbf{Living} & \textbf{Rel. $\Delta$} \\
\midrule
2WikiMultiHopQA & 1,000 & 6,038.2 & 6,241.6 & \tokinc{+3.4\%} & 1,067.6 & 906.4 & \tokdec{-15.1\%} & \$6.22 & \$5.84 & \tokdec{-6.1\%} \\
HotpotQA & 1,000 & 5,779.8 & 5,985.8 & \tokinc{+3.6\%} & 941.9 & 767.0 & \tokdec{-18.6\%} & \$5.72 & \$5.29 & \tokdec{-7.4\%} \\
MuSiQue & 1,000 & 5,888.4 & 6,092.4 & \tokinc{+3.5\%} & 2,098.8 & 1,641.7 & \tokdec{-21.8\%} & \$9.24 & \$7.97 & \tokdec{-13.7\%} \\
MuSiQue-full & 2,417 & 8,639.3 & 8,861.0 & \tokinc{+2.6\%} & 2,223.8 & 1,643.7 & \tokdec{-26.1\%} & \$26.57 & \$22.63 & \tokdec{-14.8\%} \\
WixQA & 400 & 6,633.0 & 7,412.8 & \tokinc{+11.8\%} & 1,653.8 & 1,378.1 & \tokdec{-16.7\%} & \$3.31 & \$3.14 & \tokdec{-5.3\%} \\
\midrule
Overall (weighted) & 5,817 & 7,089.7 & 7,340.9 & \tokinc{+3.5\%} & 1,744.0 & 1,347.6 & \tokdec{-22.7\%} & \$51.05 & \$44.87 & \tokdec{-12.1\%} \\
\bottomrule
\end{tabular}
\end{table*}

\begin{figure*}[t]
\centering
\includegraphics[width=\textwidth]{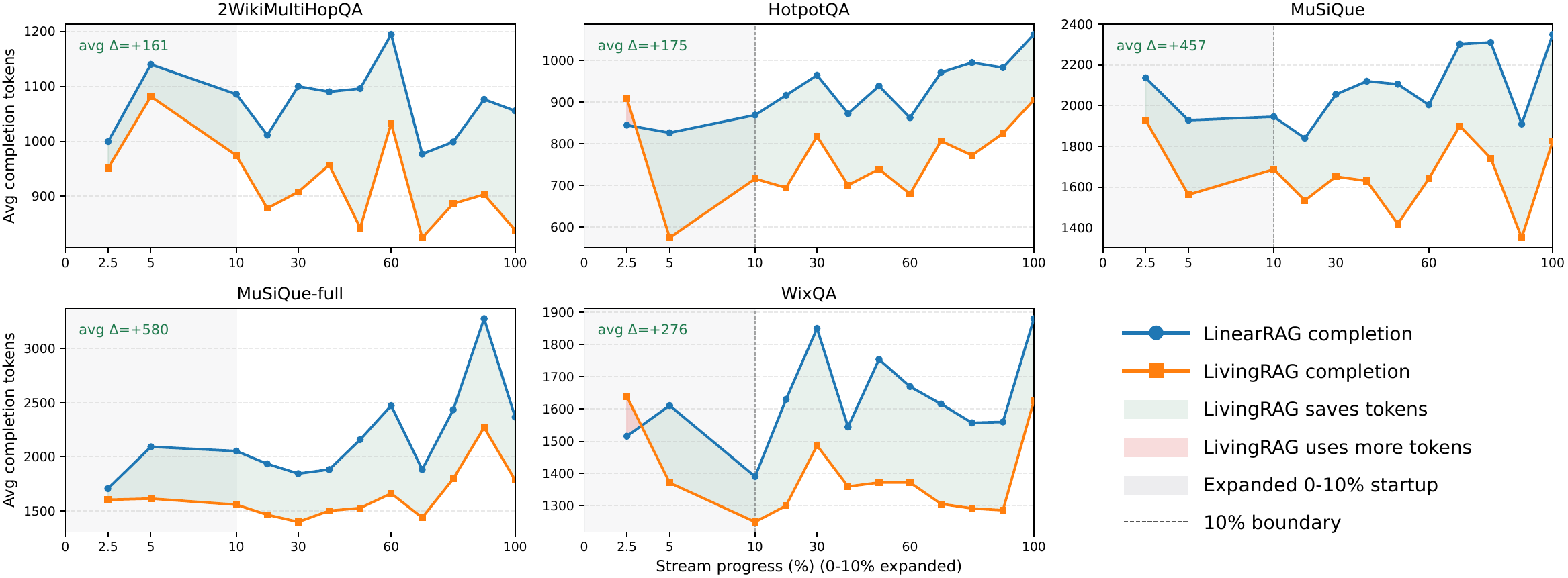}
\caption{
Completion-token trends over online QA streams.
Each panel corresponds to one dataset.
The blue and orange curves show the average completion tokens of LinearRAG and LivingRAG in each stream segment.
Green shading between the two curves marks segments where LivingRAG uses fewer completion tokens, while red shading marks segments where it uses more.
The first 10\% of each stream is expanded to show the startup stage.
}
\label{fig:completion_token_trend}
\end{figure*}

\subsection{Online Token Efficiency}
\label{sec:token-efficiency}

We next study whether online experience reuse reduces token use and estimated cost.
We use the same datasets and online setting as in Section~\ref{sec:experiments}.
Since LivingRAG uses LinearRAG as its retrieval backbone, we compare only these two systems.
This comparison isolates the effect of the writable experience store.

\paragraph{Setup.}
We record prompt tokens, completion tokens, and total run cost for each run.
Prompt tokens measure the input context.
Completion tokens measure the generated output.
Cost is estimated with the Qwen3.6 Plus prices used in Table~\ref{tab:token_efficiency}.

We report prompt tokens, completion tokens, and total run cost.
Prompt tokens correspond to the input context, while completion tokens correspond to generated output.
We report them separately because they reflect different parts of inference and are priced differently by many LLM APIs.
In our estimate, prompt tokens are charged at \$0.5 per million tokens, and completion tokens are charged at \$3.0 per million tokens.

Table~\ref{tab:token_efficiency} shows a consistent trade-off.
LivingRAG increases prompt tokens on all datasets because it may add compact reasoning scaffolds or experience-related context.
The weighted prompt-token increase is 3.5\%.
However, LivingRAG reduces completion tokens on all five datasets, with a weighted reduction of 22.7\%.
The reduction is largest on MuSiQue and MuSiQue-full, where queries often require longer multi-hop reasoning.

The total cost reduction is smaller than the completion-token reduction because the added prompt tokens partly offset the savings.
Even so, LivingRAG reduces total run cost on every dataset.
Overall, the estimated cost decreases from \$51.05 to \$44.87, a 12.1\% reduction.
Section~\ref{sec:realized-reuse-signals} further analyzes which reuse paths are used during inference.

Figure~\ref{fig:completion_token_trend} shows the online trend.
In the earliest segments, LivingRAG has few stored experiences and may temporarily use more completion tokens.
After experience accumulates, the area between the two curves is green in most segments, indicating that LivingRAG usually uses fewer completion tokens.
The segment-level trend suggests that the savings are not caused by a single late-stage average and become more stable after the startup stage.

\subsection{Realized Reuse Signals}
\label{sec:realized-reuse-signals}

We next examine whether LivingRAG uses the reuse signals that motivate our design.
This analysis is not an accuracy evaluation.
It is also not a direct conversion from reuse opportunity to realized reuse.
The opportunity values in Section~\ref{sec:motivating-analysis} describe whether a stream contains possible reuse signals.
The realized values here are measured from LivingRAG inference traces.

We measure two realized signals.
\emph{Graph-neighborhood transfer} occurs when activation-map fusion introduces new above-threshold graph entities.
\emph{Template reuse} occurs when the selected scaffold has masked-template similarity of at least 0.35 with the current query.

\begin{table}[t]
\centering
\small
\setlength{\tabcolsep}{4.2pt}
\caption{
Reuse opportunities and realized reuse in LivingRAG.
``Opp.'' denotes the template-reuse opportunity from Section~\ref{sec:motivating-analysis}.
``Reuse'' denotes realized scaffold reuse during inference.
Graph transfer is measured from activation-map fusion.
The opportunity and realized values use different measurements, so the table should be read as a trend comparison rather than a direct numerical comparison.
All values are percentages.
}
\label{tab:realized_reuse_signals}
\begin{tabular*}{\columnwidth}{@{\extracolsep{\fill}}lccc@{}}
\toprule
\multirow{2}{*}{\textbf{Dataset}}
& \multicolumn{1}{c}{\textbf{Graph signal}}
& \multicolumn{2}{c}{\textbf{Template signal}} \\
\cmidrule(lr){2-2} \cmidrule(lr){3-4}
& \textbf{Graph trans.} & \textbf{Opp.} & \textbf{Reuse} \\
\midrule
2Wiki        & 0.00  & 96.30 & 84.15 \\
HotpotQA     & 0.00  & 21.82 & 9.04  \\
MuSiQue      & 91.20 & 39.14 & 40.69 \\
MuSiQue-full & 82.42 & 54.30 & 50.44 \\
WixQA        & 91.50 & 16.04 & 7.03  \\
\bottomrule
\end{tabular*}
\end{table}

Table~\ref{tab:realized_reuse_signals} shows that LivingRAG uses different reuse paths on different datasets.
On 2Wiki, template reuse is high.
This matches the strong template signal observed in Section~\ref{sec:motivating-analysis}.
Graph transfer is 0.00\% under our strict definition.
This does not mean that graph signals are absent.
It only means that activation-map fusion does not introduce new above-threshold entities in this setting.

On MuSiQue, MuSiQue-full, and WixQA, graph transfer is much stronger.
This shows that stored activation maps often change the graph retrieval state.
Template reuse is also visible on MuSiQue and MuSiQue-full.
These results support the two reuse paths in LivingRAG.
Activation maps help retrieval reuse graph neighborhoods.
Scaffolds help generation reuse reasoning patterns.

HotpotQA has low values under both strict trace metrics.
This suggests that these two measurements may not capture all experience effects on this dataset.
Overall, the table shows that the realized behavior of LivingRAG is consistent with the reuse signals that motivated the method.
Appendix~\ref{app:additional-robustness} further analyzes performance by realized reuse signal and robustness across query orders.

\subsection{Ablation Study}
\label{sec:ablation}

We ablate three key components on 2WikiMultiHopQA, MuSiQue-full, and WixQA: activation-map fusion, reasoning scaffolds, and the write-back quality gate.
Each run starts with an empty experience store and updates it online.

\begin{table}[t]
\centering
\small
\setlength{\tabcolsep}{4pt}
\caption{Ablation results. 2W, MQ-F, and Wix report LLM accuracy (\%) on 2WikiMultiHopQA, MuSiQue-full, and WixQA. 
Red. is average completion-token reduction against LinearRAG.}
\label{tab:ablation}
\begin{tabular*}{\columnwidth}{@{\extracolsep{\fill}}lrrrr@{}}
\toprule
\textbf{Variant}
& \textbf{2W}
& \textbf{MQ-F}
& \textbf{Wix}
& \textbf{Red.} \\
\midrule
\textbf{LivingRAG}
& 87.00 & 58.42 & 70.25 & 23.6\% \\
w/o Act. fusion
& 86.70 & 56.32 & 67.45 & 15.1\% \\
w/o Scaffold
& 83.80 & 56.92 & 69.55 & 8.3\% \\
w/o Quality Gate
& 84.60 & 55.52 & 68.25 & 18.0\% \\
\bottomrule
\end{tabular*}
\end{table}

Table~\ref{tab:ablation} shows that the two experience-use paths play different roles.
Removing activation-map fusion hurts MuSiQue-full and WixQA more, where activation-map transfer is frequently realized during inference.
Removing scaffolds hurts 2WikiMultiHopQA most and also reduces completion-token savings, suggesting that scaffolds help avoid repeated reasoning.
Across the five streams, the quality gates store only 27.4\% of candidate experiences (Appendix~\ref{app:eqs-gates}).
Removing the quality gate lowers accuracy on all datasets, confirming that writable experience must pass checks before reuse.
Overall, the ablation supports the design of LivingRAG.
Activation maps help retrieval, scaffolds help generation, and quality gates keep online experience reliable.


\section{Conclusion}

We presented LivingRAG, a Graph RAG framework with writable and reusable reasoning experience.
LivingRAG stores verified experiences.
It reuses them through activation-map fusion and reasoning scaffolds.
Experiments on multi-hop QA benchmarks show that LivingRAG improves accuracy and reduces completion tokens over strong RAG baselines.
The results suggest that verified experience reuse is a simple way to augment Graph RAG.

\section{Limitations}

One limitation of LivingRAG is long-term experience growth.
As shown in our complexity analysis, LivingRAG keeps the LinearRAG graph unchanged and adds experience-store costs that are linear in the number of accepted experiences.
The runtime breakdown also shows that this overhead is acceptable in our experiments, and that post-answer validation does not need to increase user-visible latency.
However, in much larger deployments, experiences may accumulate over a long period.
Too many repeated or weakly useful experiences may increase matching cost and make reuse less stable.
Future systems may need stronger store maintenance, such as pruning, compression, or indexed experience search.

LivingRAG also assumes that stored experiences remain valid for later queries.
The current store does not update or delete accepted experiences.
The assumption can fail in dynamic domains.
For example, product policies, event information, or other time-sensitive facts may change after an experience is written.
Although each experience stores a timestamp, the current system does not explicitly reduce the influence of old experience.
The fixed-corpus benchmarks contain no chronological updates or fact-validity intervals, so they cannot directly evaluate expiration under changing knowledge.
Future work should explore time-aware refresh and adaptive pruning, so that older experiences can be downweighted, updated, or removed when new evidence appears.

Our evaluation also treats controlled QA benchmarks as proxies for online query streams.
MuSiQue is constructed by composing single-hop questions and may therefore contain denser reusable structure than naturally occurring traffic.
WixQA is more deployment-oriented, but it is not timestamped and its questions share an enterprise knowledge base.
In natural streams, reusable experiences may be more scattered, while topic drift and noise may reduce their value.
Evaluation on naturally timestamped streams with topic drift and controlled noise remains future work.

\section*{Acknowledgments}

This work was supported by the Zhejiang Provincial Natural Science Foundation of China (Youth Program) under Grant No. LQN26F020018 and by the ``Pioneer'' and ``Leading Goose'' R\&D Program of Zhejiang under Grant No. 2024C01020.

\nocite{*}
\bibliography{custom}

@article{linearRAG,
  title={LinearRAG: Linear Graph Retrieval Augmented Generation on Large-scale Corpora},
  author={Zhuang, Luyao and Chen, Shengyuan and Xiao, Yilin and Zhou, Huachi and Zhang, Yujing and Chen, Hao and Zhang, Qinggang and Huang, Xiao},
  journal={arXiv preprint arXiv:2510.10114},
  year={2025}
}

@inproceedings{hippoRAG,
    title={HippoRAG: Neurobiologically Inspired Long-Term Memory for Large Language Models}, 
    author={Bernal Jiménez Gutiérrez and Yiheng Shu and Yu Gu and Michihiro Yasunaga and Yu Su},
    booktitle={The Thirty-eighth Annual Conference on Neural Information Processing Systems},
    year={2024},
    url={https://openreview.net/forum?id=hkujvAPVsg}
}

@article{graphRAG,
  title={From local to global: A graph rag approach to query-focused summarization},
  author={Edge, Darren and Trinh, Ha and Cheng, Newman and Bradley, Joshua and Chao, Alex and Mody, Apurva and Truitt, Steven and Metropolitansky, Dasha and Ness, Robert Osazuwa and Larson, Jonathan},
  journal={arXiv preprint arXiv:2404.16130},
  year={2024}
}

@article{lightRAG,
  title={Lightrag: Simple and fast retrieval-augmented generation},
  author={Guo, Zirui and Xia, Lianghao and Yu, Yanhua and Ao, Tian and Huang, Chao},
  journal={arXiv preprint arXiv:2410.05779},
  volume={2},
  number={3},
  year={2024}
}

@inproceedings{
ma2025thinkongraph,
title={Think-on-Graph 2.0: Deep and Faithful Large Language Model Reasoning with Knowledge-guided Retrieval Augmented Generation},
author={Shengjie Ma and Chengjin Xu and Xuhui Jiang and Muzhi Li and Huaren Qu and Cehao Yang and Jiaxin Mao and Jian Guo},
booktitle={The Thirteenth International Conference on Learning Representations},
year={2025},
url={https://openreview.net/forum?id=oFBu7qaZpS}
}

@inproceedings{
li2025simple,
title={Simple is Effective: The Roles of Graphs and Large Language Models in Knowledge-Graph-Based Retrieval-Augmented Generation},
author={Mufei Li and Siqi Miao and Pan Li},
booktitle={ICLR 2025 Workshop on Foundation Models in the Wild},
year={2025},
url={https://openreview.net/forum?id=2NbxnNI94F}
}

@inproceedings{
hippoRAG2,
title={From {RAG} to Memory: Non-Parametric Continual Learning for Large Language Models},
author={Bernal Jim{\'e}nez Guti{\'e}rrez and Yiheng Shu and Weijian Qi and Sizhe Zhou and Yu Su},
booktitle={Forty-second International Conference on Machine Learning},
year={2025},
url={https://openreview.net/forum?id=LWH8yn4HS2}
}

@inproceedings{
gfmRAG,
title={{GFM}-{RAG}: Graph Foundation Model for Retrieval Augmented Generation},
author={Linhao Luo and Zicheng Zhao and Gholamreza Haffari and Dinh Phung and Chen Gong and Shirui Pan},
booktitle={The Thirty-ninth Annual Conference on Neural Information Processing Systems},
year={2026},
url={https://openreview.net/forum?id=0QNmAvQQqj}
}

@inproceedings{
luo2026hypergraphrag,
title={HyperGraph{RAG}: Retrieval-Augmented Generation via Hypergraph-Structured Knowledge Representation},
author={Haoran Luo and Haihong E and Guanting Chen and Yandan Zheng and Xiaobao Wu and Yikai Guo and Qika Lin and Yu Feng and Zemin Kuang and Meina Song and Yifan Zhu and Anh Tuan Luu},
booktitle={The Thirty-ninth Annual Conference on Neural Information Processing Systems},
year={2026},
url={https://openreview.net/forum?id=ravS5h8MNg}
}

@inproceedings{
graphFlow,
title={Can Knowledge-Graph-based Retrieval Augmented Generation Really Retrieve What You Need?},
author={Junchi Yu and Yujie Liu and Jindong Gu and Philip Torr and Dongzhan Zhou},
booktitle={The Thirty-ninth Annual Conference on Neural Information Processing Systems},
year={2026},
url={https://openreview.net/forum?id=po0eyoYFUa}
}

@inproceedings{
he2024gretriever,
title={G-Retriever: Retrieval-Augmented Generation for Textual Graph Understanding and Question Answering},
author={Xiaoxin He and Yijun Tian and Yifei Sun and Nitesh V Chawla and Thomas Laurent and Yann LeCun and Xavier Bresson and Bryan Hooi},
booktitle={The Thirty-eighth Annual Conference on Neural Information Processing Systems},
year={2024},
url={https://openreview.net/forum?id=MPJ3oXtTZl}
}

@inproceedings{
luo2024reasoning,
title={Reasoning on Graphs: Faithful and Interpretable Large Language Model Reasoning},
author={LINHAO LUO and Yuan-Fang Li and Gholamreza Haffari and Shirui Pan},
booktitle={The Twelfth International Conference on Learning Representations},
year={2024},
url={https://openreview.net/forum?id=ZGNWW7xZ6Q}
}

@inproceedings{
xiang2026when,
title={When to use Graphs in {RAG}: A Comprehensive Analysis for Graph Retrieval-Augmented Generation},
author={Zhishang Xiang and Chuanjie Wu and Qinggang Zhang and Shengyuan Chen and Zijin Hong and Xiao Huang and Jinsong Su},
booktitle={The Fourteenth International Conference on Learning Representations},
year={2026},
url={https://openreview.net/forum?id=i9q9xDMjG7}
}

@inproceedings{
shu2026remem,
title={{REM}em: Reasoning with Episodic Memory in Language Agent},
author={Yiheng Shu and Saisri Padmaja Jonnalagedda and Xiang Gao and Bernal Jim{\'e}nez Guti{\'e}rrez and Weijian Qi and Kamalika Das and Huan Sun and Yu Su},
booktitle={The Fourteenth International Conference on Learning Representations},
year={2026},
url={https://openreview.net/forum?id=fugnQxbvMm}
}

@inproceedings{
zhou2026queryaware,
title={Query-Aware Flow Diffusion for Graph-Based {RAG} with Retrieval Guarantees},
author={Zhuoping Zhou and Davoud Ataee Tarzanagh and Sima Didari and Wenjun Hu and Baruch Gutow and Oxana Verkholyak and Masoud Faraki and Heng Hao and Hankyu Moon and Seungjai Min},
booktitle={The Fourteenth International Conference on Learning Representations},
year={2026},
url={https://openreview.net/forum?id=n28wnc2QTc}
}

@misc{dong2025youtugraphrag,
      title={Youtu-GraphRAG: Vertically Unified Agents for Graph Retrieval-Augmented Complex Reasoning}, 
      author={Junnan Dong and Siyu An and Yifei Yu and Qian-Wen Zhang and Linhao Luo and Xiao Huang and Yunsheng Wu and Di Yin and Xing Sun},
      year={2025},
      eprint={2508.19855},
      archivePrefix={arXiv},
      url={https://arxiv.org/abs/2508.19855}, 
}

@inproceedings{gao-etal-2025-frag,
    title = "{FRAG}: A Flexible Modular Framework for Retrieval-Augmented Generation based on Knowledge Graphs",
    author = "Gao, Zengyi  and
      Cao, Yukun  and
      Wang, Hairu  and
      Ke, Ao  and
      Feng, Yuan  and
      Zhou, S Kevin  and
      Xie, Xike",
    editor = "Che, Wanxiang  and
      Nabende, Joyce  and
      Shutova, Ekaterina  and
      Pilehvar, Mohammad Taher",
    booktitle = "Findings of the Association for Computational Linguistics: ACL 2025",
    month = jul,
    year = "2025",
    address = "Vienna, Austria",
    publisher = "Association for Computational Linguistics",
    url = "https://aclanthology.org/2025.findings-acl.321/",
    doi = "10.18653/v1/2025.findings-acl.321",
    pages = "6178--6192",
    ISBN = "979-8-89176-256-5"
}

@inproceedings{cai-etal-2025-simgrag,
    title = "{S}im{GRAG}: Leveraging Similar Subgraphs for Knowledge Graphs Driven Retrieval-Augmented Generation",
    author = "Cai, Yuzheng  and
      Guo, Zhenyue  and
      Pei, YiWen  and
      Bian, WanRui  and
      Zheng, Weiguo",
    editor = "Che, Wanxiang  and
      Nabende, Joyce  and
      Shutova, Ekaterina  and
      Pilehvar, Mohammad Taher",
    booktitle = "Findings of the Association for Computational Linguistics: ACL 2025",
    month = jul,
    year = "2025",
    address = "Vienna, Austria",
    publisher = "Association for Computational Linguistics",
    url = "https://aclanthology.org/2025.findings-acl.163/",
    doi = "10.18653/v1/2025.findings-acl.163",
    pages = "3139--3158",
    ISBN = "979-8-89176-256-5"
}

@inproceedings{liu-etal-2025-hoprag,
    title = "{H}op{RAG}: Multi-Hop Reasoning for Logic-Aware Retrieval-Augmented Generation",
    author = "Liu, Hao  and
      Wang, Zhengren  and
      Chen, Xi  and
      Li, Zhiyu  and
      Xiong, Feiyu  and
      Yu, Qinhan  and
      Zhang, Wentao",
    editor = "Che, Wanxiang  and
      Nabende, Joyce  and
      Shutova, Ekaterina  and
      Pilehvar, Mohammad Taher",
    booktitle = "Findings of the Association for Computational Linguistics: ACL 2025",
    month = jul,
    year = "2025",
    address = "Vienna, Austria",
    publisher = "Association for Computational Linguistics",
    url = "https://aclanthology.org/2025.findings-acl.97/",
    doi = "10.18653/v1/2025.findings-acl.97",
    pages = "1897--1913",
    ISBN = "979-8-89176-256-5"
}

@inproceedings{shen-etal-2025-gear,
    title = "{G}e{AR}: Graph-enhanced Agent for Retrieval-augmented Generation",
    author = "Shen, Zhili  and
      Diao, Chenxin  and
      Vougiouklis, Pavlos  and
      Merita, Pascual  and
      Piramanayagam, Shriram  and
      Chen, Enting  and
      Graux, Damien  and
      Melo, Andre  and
      Lai, Ruofei  and
      Jiang, Zeren  and
      Li, Zhongyang  and
      Qi, Ye  and
      Ren, Yang  and
      Tu, Dandan  and
      Pan, Jeff Z.",
    editor = "Che, Wanxiang  and
      Nabende, Joyce  and
      Shutova, Ekaterina  and
      Pilehvar, Mohammad Taher",
    booktitle = "Findings of the Association for Computational Linguistics: ACL 2025",
    month = jul,
    year = "2025",
    address = "Vienna, Austria",
    publisher = "Association for Computational Linguistics",
    url = "https://aclanthology.org/2025.findings-acl.624/",
    doi = "10.18653/v1/2025.findings-acl.624",
    pages = "12049--12072",
    ISBN = "979-8-89176-256-5"
}

@inproceedings{lee-etal-2025-hybgrag,
    title = "{H}yb{GRAG}: Hybrid Retrieval-Augmented Generation on Textual and Relational Knowledge Bases",
    author = "Lee, Meng-Chieh  and
      Zhu, Qi  and
      Mavromatis, Costas  and
      Han, Zhen  and
      Adeshina, Soji  and
      Ioannidis, Vassilis N.  and
      Rangwala, Huzefa  and
      Faloutsos, Christos",
    editor = "Che, Wanxiang  and
      Nabende, Joyce  and
      Shutova, Ekaterina  and
      Pilehvar, Mohammad Taher",
    booktitle = "Proceedings of the 63rd Annual Meeting of the Association for Computational Linguistics (Volume 1: Long Papers)",
    month = jul,
    year = "2025",
    address = "Vienna, Austria",
    publisher = "Association for Computational Linguistics",
    url = "https://aclanthology.org/2025.acl-long.43/",
    doi = "10.18653/v1/2025.acl-long.43",
    pages = "879--893",
    ISBN = "979-8-89176-251-0"
}

@inproceedings{jiang-etal-2025-hykge,
    title = "{H}y{KGE}: A Hypothesis Knowledge Graph Enhanced {RAG} Framework for Accurate and Reliable Medical {LLM}s Responses",
    author = "Jiang, Xinke  and
      Zhang, Ruizhe  and
      Xu, Yongxin  and
      Qiu, Rihong  and
      Fang, Yue  and
      Wang, Zhiyuan  and
      Tang, Jinyi  and
      Ding, Hongxin  and
      Chu, Xu  and
      Zhao, Junfeng  and
      Wang, Yasha",
    editor = "Che, Wanxiang  and
      Nabende, Joyce  and
      Shutova, Ekaterina  and
      Pilehvar, Mohammad Taher",
    booktitle = "Proceedings of the 63rd Annual Meeting of the Association for Computational Linguistics (Volume 1: Long Papers)",
    month = jul,
    year = "2025",
    address = "Vienna, Austria",
    publisher = "Association for Computational Linguistics",
    url = "https://aclanthology.org/2025.acl-long.580/",
    doi = "10.18653/v1/2025.acl-long.580",
    pages = "11836--11856",
    ISBN = "979-8-89176-251-0"
}

@inproceedings{wang-etal-2025-causalrag,
    title = "{C}ausal{RAG}: Integrating Causal Graphs into Retrieval-Augmented Generation",
    author = "Wang, Nengbo  and
      Han, Xiaotian  and
      Singh, Jagdip  and
      Ma, Jing  and
      Chaudhary, Vipin",
    editor = "Che, Wanxiang  and
      Nabende, Joyce  and
      Shutova, Ekaterina  and
      Pilehvar, Mohammad Taher",
    booktitle = "Findings of the Association for Computational Linguistics: ACL 2025",
    month = jul,
    year = "2025",
    address = "Vienna, Austria",
    publisher = "Association for Computational Linguistics",
    url = "https://aclanthology.org/2025.findings-acl.1165/",
    doi = "10.18653/v1/2025.findings-acl.1165",
    pages = "22680--22693",
    ISBN = "979-8-89176-256-5"
}

@inproceedings{fang-etal-2025-kirag,
    title = "{K}i{RAG}: Knowledge-Driven Iterative Retriever for Enhancing Retrieval-Augmented Generation",
    author = "Fang, Jinyuan  and
      Meng, Zaiqiao  and
      MacDonald, Craig",
    editor = "Che, Wanxiang  and
      Nabende, Joyce  and
      Shutova, Ekaterina  and
      Pilehvar, Mohammad Taher",
    booktitle = "Proceedings of the 63rd Annual Meeting of the Association for Computational Linguistics (Volume 1: Long Papers)",
    month = jul,
    year = "2025",
    address = "Vienna, Austria",
    publisher = "Association for Computational Linguistics",
    url = "https://aclanthology.org/2025.acl-long.929/",
    doi = "10.18653/v1/2025.acl-long.929",
    pages = "18969--18985",
    ISBN = "979-8-89176-251-0"
}

@inproceedings{zhang-etal-2025-graph,
    title = "Graph of Records: Boosting Retrieval Augmented Generation for Long-context Summarization with Graphs",
    author = "Zhang, Haozhen  and
      Feng, Tao  and
      You, Jiaxuan",
    editor = "Che, Wanxiang  and
      Nabende, Joyce  and
      Shutova, Ekaterina  and
      Pilehvar, Mohammad Taher",
    booktitle = "Proceedings of the 63rd Annual Meeting of the Association for Computational Linguistics (Volume 1: Long Papers)",
    month = jul,
    year = "2025",
    address = "Vienna, Austria",
    publisher = "Association for Computational Linguistics",
    url = "https://aclanthology.org/2025.acl-long.1159/",
    doi = "10.18653/v1/2025.acl-long.1159",
    pages = "23780--23799",
    ISBN = "979-8-89176-251-0"
}

@inproceedings{long-etal-2025-drae,
    title = "{DRAE}: Dynamic Retrieval-Augmented Expert Networks for Lifelong Learning and Task Adaptation in Robotics",
    author = "Long, Yayu  and
      Chen, Kewei  and
      Jin, Long  and
      Shang, Mingsheng",
    editor = "Che, Wanxiang  and
      Nabende, Joyce  and
      Shutova, Ekaterina  and
      Pilehvar, Mohammad Taher",
    booktitle = "Proceedings of the 63rd Annual Meeting of the Association for Computational Linguistics (Volume 1: Long Papers)",
    month = jul,
    year = "2025",
    address = "Vienna, Austria",
    publisher = "Association for Computational Linguistics",
    url = "https://aclanthology.org/2025.acl-long.1127/",
    doi = "10.18653/v1/2025.acl-long.1127",
    pages = "23098--23141",
    ISBN = "979-8-89176-251-0"
}

@inproceedings{guo-etal-2025-empowering,
    title = "Empowering {G}raph{RAG} with Knowledge Filtering and Integration",
    author = "Guo, Kai  and
      Shomer, Harry  and
      Zeng, Shenglai  and
      Han, Haoyu  and
      Wang, Yu  and
      Tang, Jiliang",
    editor = "Christodoulopoulos, Christos  and
      Chakraborty, Tanmoy  and
      Rose, Carolyn  and
      Peng, Violet",
    booktitle = "Proceedings of the 2025 Conference on Empirical Methods in Natural Language Processing",
    month = nov,
    year = "2025",
    address = "Suzhou, China",
    publisher = "Association for Computational Linguistics",
    url = "https://aclanthology.org/2025.emnlp-main.1293/",
    doi = "10.18653/v1/2025.emnlp-main.1293",
    pages = "25439--25453",
    ISBN = "979-8-89176-332-6"
}

@inproceedings{mavromatis-etal-2025-byokg,
    title = "{BYOKG}-{RAG}: Multi-Strategy Graph Retrieval for Knowledge Graph Question Answering",
    author = "Mavromatis, Costas  and
      Adeshina, Soji  and
      Ioannidis, Vassilis N.  and
      Han, Zhen  and
      Zhu, Qi  and
      Robinson, Ian  and
      Thompson, Bryan  and
      Rangwala, Huzefa  and
      Karypis, George",
    editor = "Christodoulopoulos, Christos  and
      Chakraborty, Tanmoy  and
      Rose, Carolyn  and
      Peng, Violet",
    booktitle = "Proceedings of the 2025 Conference on Empirical Methods in Natural Language Processing",
    month = nov,
    year = "2025",
    address = "Suzhou, China",
    publisher = "Association for Computational Linguistics",
    url = "https://aclanthology.org/2025.emnlp-main.1417/",
    doi = "10.18653/v1/2025.emnlp-main.1417",
    pages = "27881--27898",
    ISBN = "979-8-89176-332-6"
}

@inproceedings{gao-etal-2025-rag,
    title = "{D}-{RAG}: Differentiable Retrieval-Augmented Generation for Knowledge Graph Question Answering",
    author = "Gao, Guangze  and
      Li, Zixuan  and
      Yuan, Chunfeng  and
      Li, Jiawei  and
      Jianzhuo, Wu  and
      Zhang, Yuehao  and
      Jin, Xiaolong  and
      Li, Bing  and
      Hu, Weiming",
    editor = "Christodoulopoulos, Christos  and
      Chakraborty, Tanmoy  and
      Rose, Carolyn  and
      Peng, Violet",
    booktitle = "Proceedings of the 2025 Conference on Empirical Methods in Natural Language Processing",
    month = nov,
    year = "2025",
    address = "Suzhou, China",
    publisher = "Association for Computational Linguistics",
    url = "https://aclanthology.org/2025.emnlp-main.1793/",
    doi = "10.18653/v1/2025.emnlp-main.1793",
    pages = "35398--35417",
    ISBN = "979-8-89176-332-6"
}

@inproceedings{helali-etal-2025-reliable,
    title = "Reliable and Cost-Effective Exploratory Data Analysis via Graph-Guided {RAG}",
    author = "Helali, Mossad  and
      Luo, Yutai  and
      Ham, Tae Jun  and
      Plotts, Jim  and
      Chaugule, Ashwin  and
      Chang, Jichuan  and
      Ranganathan, Parthasarathy  and
      Mansour, Essam",
    editor = "Christodoulopoulos, Christos  and
      Chakraborty, Tanmoy  and
      Rose, Carolyn  and
      Peng, Violet",
    booktitle = "Proceedings of the 2025 Conference on Empirical Methods in Natural Language Processing",
    month = nov,
    year = "2025",
    address = "Suzhou, China",
    publisher = "Association for Computational Linguistics",
    url = "https://aclanthology.org/2025.emnlp-main.836/",
    doi = "10.18653/v1/2025.emnlp-main.836",
    pages = "16536--16553",
    ISBN = "979-8-89176-332-6"
}

@inproceedings{guan-etal-2025-kg,
    title = "{KG}-{RAG}: Enhancing {GUI} Agent Decision-Making via Knowledge Graph-Driven Retrieval-Augmented Generation",
    author = "Guan, Ziyi  and
      Li, Jason Chun Lok  and
      Hou, Zhijian  and
      Zhang, Pingping  and
      Xu, Donglai  and
      Zhao, Yuzhi  and
      Wu, Mengyang  and
      Chen, Jinpeng  and
      Nguyen, Thanh-Toan  and
      Xian, Pengfei  and
      Ma, Wenao  and
      Qin, Shengchao  and
      Chesi, Graziano  and
      Wong, Ngai",
    editor = "Christodoulopoulos, Christos  and
      Chakraborty, Tanmoy  and
      Rose, Carolyn  and
      Peng, Violet",
    booktitle = "Proceedings of the 2025 Conference on Empirical Methods in Natural Language Processing",
    month = nov,
    year = "2025",
    address = "Suzhou, China",
    publisher = "Association for Computational Linguistics",
    url = "https://aclanthology.org/2025.emnlp-main.274/",
    doi = "10.18653/v1/2025.emnlp-main.274",
    pages = "5396--5405",
    ISBN = "979-8-89176-332-6"
}

@inproceedings{wang-han-2025-proprag,
    title = "{P}rop{RAG}: Guiding Retrieval with Beam Search over Proposition Paths",
    author = "Wang, Jingjin  and
      Han, Jiawei",
    editor = "Christodoulopoulos, Christos  and
      Chakraborty, Tanmoy  and
      Rose, Carolyn  and
      Peng, Violet",
    booktitle = "Proceedings of the 2025 Conference on Empirical Methods in Natural Language Processing",
    month = nov,
    year = "2025",
    address = "Suzhou, China",
    publisher = "Association for Computational Linguistics",
    url = "https://aclanthology.org/2025.emnlp-main.317/",
    doi = "10.18653/v1/2025.emnlp-main.317",
    pages = "6212--6227",
    ISBN = "979-8-89176-332-6"
}

@inproceedings{zhang-etal-2025-pathwiserag,
    title = "{P}athwise{RAG}: Multi-Dimensional Exploration and Integration Framework",
    author = "Zhang, Hengrui  and
      Huang, Pin-Siang  and
      Zhang, Zhen  and
      Lin, Peican  and
      Yu, Yao-Ching  and
      Hu, Bo  and
      Du, Yulu",
    editor = "Christodoulopoulos, Christos  and
      Chakraborty, Tanmoy  and
      Rose, Carolyn  and
      Peng, Violet",
    booktitle = "Proceedings of the 2025 Conference on Empirical Methods in Natural Language Processing",
    month = nov,
    year = "2025",
    address = "Suzhou, China",
    publisher = "Association for Computational Linguistics",
    url = "https://aclanthology.org/2025.emnlp-main.1167/",
    doi = "10.18653/v1/2025.emnlp-main.1167",
    pages = "22904--22925",
    ISBN = "979-8-89176-332-6"
}

@inproceedings{tan-etal-2025-hydrarag,
    title = "{H}ydra{RAG}: Structured Cross-Source Enhanced Large Language Model Reasoning",
    author = "Tan, Xingyu  and
      Wang, Xiaoyang  and
      Liu, Qing  and
      Xu, Xiwei  and
      Yuan, Xin  and
      Zhu, Liming  and
      Zhang, Wenjie",
    editor = "Christodoulopoulos, Christos  and
      Chakraborty, Tanmoy  and
      Rose, Carolyn  and
      Peng, Violet",
    booktitle = "Proceedings of the 2025 Conference on Empirical Methods in Natural Language Processing",
    month = nov,
    year = "2025",
    address = "Suzhou, China",
    publisher = "Association for Computational Linguistics",
    url = "https://aclanthology.org/2025.emnlp-main.730/",
    doi = "10.18653/v1/2025.emnlp-main.730",
    pages = "14431--14459",
    ISBN = "979-8-89176-332-6"
}

@inproceedings{wang-etal-2025-infogain,
    title = "{I}nfo{G}ain-{RAG}: Boosting Retrieval-Augmented Generation through Document Information Gain-based Reranking and Filtering",
    author = "Wang, Zihan  and
      Liang, Zihan  and
      Shao, Zhou  and
      Ma, Yufei  and
      Dai, Huangyu  and
      Chen, Ben  and
      Mao, Lingtao  and
      Lei, Chenyi  and
      Ding, Yuqing  and
      Li, Han",
    editor = "Christodoulopoulos, Christos  and
      Chakraborty, Tanmoy  and
      Rose, Carolyn  and
      Peng, Violet",
    booktitle = "Proceedings of the 2025 Conference on Empirical Methods in Natural Language Processing",
    month = nov,
    year = "2025",
    address = "Suzhou, China",
    publisher = "Association for Computational Linguistics",
    url = "https://aclanthology.org/2025.emnlp-main.365/",
    doi = "10.18653/v1/2025.emnlp-main.365",
    pages = "7190--7204",
    ISBN = "979-8-89176-332-6"
}

@inproceedings{xu-etal-2025-training,
    title = "Training a Utility-based Retriever Through Shared Context Attribution for Retrieval-Augmented Language Models",
    author = "Xu, Yilong  and
      Gao, Jinhua  and
      Yu, Xiaoming  and
      Xue, Yuanhai  and
      Bi, Baolong  and
      Shen, Huawei  and
      Cheng, Xueqi",
    editor = "Christodoulopoulos, Christos  and
      Chakraborty, Tanmoy  and
      Rose, Carolyn  and
      Peng, Violet",
    booktitle = "Proceedings of the 2025 Conference on Empirical Methods in Natural Language Processing",
    month = nov,
    year = "2025",
    address = "Suzhou, China",
    publisher = "Association for Computational Linguistics",
    url = "https://aclanthology.org/2025.emnlp-main.33/",
    doi = "10.18653/v1/2025.emnlp-main.33",
    pages = "629--648",
    ISBN = "979-8-89176-332-6"
}

@inproceedings{salama-etal-2025-meminsight,
    title = "{M}em{I}nsight: Autonomous Memory Augmentation for {LLM} Agents",
    author = "Salama, Rana  and
      Cai, Jason  and
      Yuan, Michelle  and
      Currey, Anna  and
      Sunkara, Monica  and
      Zhang, Yi  and
      Benajiba, Yassine",
    editor = "Christodoulopoulos, Christos  and
      Chakraborty, Tanmoy  and
      Rose, Carolyn  and
      Peng, Violet",
    booktitle = "Proceedings of the 2025 Conference on Empirical Methods in Natural Language Processing",
    month = nov,
    year = "2025",
    address = "Suzhou, China",
    publisher = "Association for Computational Linguistics",
    url = "https://aclanthology.org/2025.emnlp-main.1683/",
    doi = "10.18653/v1/2025.emnlp-main.1683",
    pages = "33136--33152",
    ISBN = "979-8-89176-332-6"
}

@inproceedings{kang-etal-2025-memory,
    title = "Memory {OS} of {AI} Agent",
    author = "Kang, Jiazheng  and
      Ji, Mingming  and
      Zhao, Zhe  and
      Bai, Ting",
    editor = "Christodoulopoulos, Christos  and
      Chakraborty, Tanmoy  and
      Rose, Carolyn  and
      Peng, Violet",
    booktitle = "Proceedings of the 2025 Conference on Empirical Methods in Natural Language Processing",
    month = nov,
    year = "2025",
    address = "Suzhou, China",
    publisher = "Association for Computational Linguistics",
    url = "https://aclanthology.org/2025.emnlp-main.1318/",
    doi = "10.18653/v1/2025.emnlp-main.1318",
    pages = "25961--25970",
    ISBN = "979-8-89176-332-6"
}

@misc{gao2023retrieval,
      title={Retrieval-Augmented Generation for Large Language Models: A Survey}, 
      author={Yunfan Gao and Yun Xiong and Xinyu Gao and Kangxiang Jia and Jinliu Pan and Yuxi Bi and Yi Dai and Jiawei Sun and Meng Wang and Haofen Wang},
      year={2024},
      eprint={2312.10997},
      archivePrefix={arXiv},
      primaryClass={cs.CL},
      url={https://arxiv.org/abs/2312.10997}, 
}

@inproceedings{karpukhin2020dense,
    title = "Dense Passage Retrieval for Open-Domain Question Answering",
    author = "Karpukhin, Vladimir  and
      Oguz, Barlas  and
      Min, Sewon  and
      Lewis, Patrick  and
      Wu, Ledell  and
      Edunov, Sergey  and
      Chen, Danqi  and
      Yih, Wen-tau",
    editor = "Webber, Bonnie  and
      Cohn, Trevor  and
      He, Yulan  and
      Liu, Yang",
    booktitle = "Proceedings of the 2020 Conference on Empirical Methods in Natural Language Processing (EMNLP)",
    month = nov,
    year = "2020",
    address = "Online",
    publisher = "Association for Computational Linguistics",
    url = "https://aclanthology.org/2020.emnlp-main.550/",
    doi = "10.18653/v1/2020.emnlp-main.550",
    pages = "6769--6781"
}

@inproceedings{trivedi2023interleaving,
    title = "Interleaving Retrieval with Chain-of-Thought Reasoning for Knowledge-Intensive Multi-Step Questions",
    author = "Trivedi, Harsh  and
      Balasubramanian, Niranjan  and
      Khot, Tushar  and
      Sabharwal, Ashish",
    editor = "Rogers, Anna  and
      Boyd-Graber, Jordan  and
      Okazaki, Naoaki",
    booktitle = "Proceedings of the 61st Annual Meeting of the Association for Computational Linguistics (Volume 1: Long Papers)",
    month = jul,
    year = "2023",
    address = "Toronto, Canada",
    publisher = "Association for Computational Linguistics",
    url = "https://aclanthology.org/2023.acl-long.557/",
    doi = "10.18653/v1/2023.acl-long.557",
    pages = "10014--10037"
}

@inproceedings{jiang2023active,
    title = "Active Retrieval Augmented Generation",
    author = "Jiang, Zhengbao  and
      Xu, Frank  and
      Gao, Luyu  and
      Sun, Zhiqing  and
      Liu, Qian  and
      Dwivedi-Yu, Jane  and
      Yang, Yiming  and
      Callan, Jamie  and
      Neubig, Graham",
    editor = "Bouamor, Houda  and
      Pino, Juan  and
      Bali, Kalika",
    booktitle = "Proceedings of the 2023 Conference on Empirical Methods in Natural Language Processing",
    month = dec,
    year = "2023",
    address = "Singapore",
    publisher = "Association for Computational Linguistics",
    url = "https://aclanthology.org/2023.emnlp-main.495/",
    doi = "10.18653/v1/2023.emnlp-main.495",
    pages = "7969--7992"
}

@inproceedings{
wei2022chain,
title={Chain of Thought Prompting Elicits Reasoning in Large Language Models},
author={Jason Wei and Xuezhi Wang and Dale Schuurmans and Maarten Bosma and brian ichter and Fei Xia and Ed H. Chi and Quoc V Le and Denny Zhou},
booktitle={Advances in Neural Information Processing Systems},
editor={Alice H. Oh and Alekh Agarwal and Danielle Belgrave and Kyunghyun Cho},
year={2022},
url={https://openreview.net/forum?id=_VjQlMeSB_J}
}

@inproceedings{yao2022react,
  title = {{ReAct}: Synergizing Reasoning and Acting in Language Models},
  author = {Yao, Shunyu and Zhao, Jeffrey and Yu, Dian and Du, Nan and Shafran, Izhak and Narasimhan, Karthik and Cao, Yuan},
  booktitle = {International Conference on Learning Representations (ICLR) },
  year = {2023},
  html = {https://arxiv.org/abs/2210.03629},
}

@inproceedings{asai2024selfRAG,
title={Self-{RAG}: Learning to Retrieve, Generate, and Critique through Self-Reflection},
author={Akari Asai and Zeqiu Wu and Yizhong Wang and Avirup Sil and Hannaneh Hajishirzi},
booktitle={The Twelfth International Conference on Learning Representations},
year={2024},
url={https://openreview.net/forum?id=hSyW5go0v8}
}

@inproceedings{chainofNote,
    title = "Chain-of-Note: Enhancing Robustness in Retrieval-Augmented Language Models",
    author = "Yu, Wenhao  and
      Zhang, Hongming  and
      Pan, Xiaoman  and
      Cao, Peixin  and
      Ma, Kaixin  and
      Li, Jian  and
      Wang, Hongwei  and
      Yu, Dong",
    editor = "Al-Onaizan, Yaser  and
      Bansal, Mohit  and
      Chen, Yun-Nung",
    booktitle = "Proceedings of the 2024 Conference on Empirical Methods in Natural Language Processing",
    month = nov,
    year = "2024",
    address = "Miami, Florida, USA",
    publisher = "Association for Computational Linguistics",
    url = "https://aclanthology.org/2024.emnlp-main.813/",
    doi = "10.18653/v1/2024.emnlp-main.813",
    pages = "14672--14685"
}

@inproceedings{logicRAG,
  title={Logic-RAG: Augmenting Large Multimodal Models with Visual-Spatial Knowledge for Road Scene Understanding},
  author={Kabir, Imran and Reza, Md Alimoor and Billah, Syed},
  booktitle={2025 IEEE International Conference on Robotics and Automation (ICRA)},
  year={2025},
  organization={IEEE}
}

@article{continualLLMsurvey,
  title={Continual Learning of Large Language Models: A Comprehensive Survey},
  author={Shi, Haizhou and 
          Xu, Zihao and 
          Wang, Hengyi and 
          Qin, Weiyi and 
          Wang, Wenyuan and 
          Wang, Yibin and 
          Wang, Zifeng and 
          Ebrahimi, Sayna and 
          Wang, Hao},
  journal={arXiv preprint arXiv:2404.16789},
  year={2024}
}

@inproceedings{reflexion,
title={Reflexion: language agents with verbal reinforcement learning},
author={Noah Shinn and Federico Cassano and Ashwin Gopinath and Karthik R Narasimhan and Shunyu Yao},
booktitle={Thirty-seventh Conference on Neural Information Processing Systems},
year={2023},
url={https://openreview.net/forum?id=vAElhFcKW6}
}

@misc{selfRefine,
      title={Self-Refine: Iterative Refinement with Self-Feedback}, 
      author={Aman Madaan and Niket Tandon and Prakhar Gupta and Skyler Hallinan and Luyu Gao and Sarah Wiegreffe and Uri Alon and Nouha Dziri and Shrimai Prabhumoye and Yiming Yang and Sean Welleck and Bodhisattwa Prasad Majumder and Shashank Gupta and Amir Yazdanbakhsh and Peter Clark},
      year={2023},
      eprint={2303.17651},
      archivePrefix={arXiv},
      primaryClass={cs.CL}
}

@inproceedings{lewis2020retrieval,
author = {Lewis, Patrick and Perez, Ethan and Piktus, Aleksandra and Petroni, Fabio and Karpukhin, Vladimir and Goyal, Naman and K\"{u}ttler, Heinrich and Lewis, Mike and Yih, Wen-tau and Rockt\"{a}schel, Tim and Riedel, Sebastian and Kiela, Douwe},
title = {Retrieval-augmented generation for knowledge-intensive NLP tasks},
year = {2020},
isbn = {9781713829546},
publisher = {Curran Associates Inc.},
address = {Red Hook, NY, USA},
booktitle = {Proceedings of the 34th International Conference on Neural Information Processing Systems},
articleno = {793},
numpages = {16},
location = {Vancouver, BC, Canada},
series = {NIPS '20}
}

@inproceedings{MemoRAG,
author = {Qian, Hongjin and Liu, Zheng and Zhang, Peitian and Mao, Kelong and Lian, Defu and Dou, Zhicheng and Huang, Tiejun},
title = {MemoRAG: Boosting Long Context Processing with Global Memory-Enhanced Retrieval Augmentation},
year = {2025},
isbn = {9798400712746},
publisher = {Association for Computing Machinery},
address = {New York, NY, USA},
url = {https://doi.org/10.1145/3696410.3714805},
doi = {10.1145/3696410.3714805},
booktitle = {Proceedings of the ACM on Web Conference 2025},
pages = {2366–2377},
numpages = {12},
location = {Sydney NSW, Australia},
series = {WWW '25}
}

\appendix

\section{Additional Related Work}
\label{app:related-work}

\paragraph{Graph-based RAG.}
Graph-based RAG organizes external knowledge with explicit graph structures.
GraphRAG builds an entity graph and community summaries to support global sensemaking over a corpus~\citep{graphRAG}.
LightRAG combines graph indexing with vector retrieval to retrieve entities, relations, and chunks efficiently~\citep{lightRAG}.
HippoRAG and HippoRAG2 use Personalized PageRank over an automatically constructed graph to support associative and multi-hop retrieval~\citep{hippoRAG,hippoRAG2}.
LinearRAG builds a lightweight passage-sentence-entity graph and performs sparse graph propagation for efficient retrieval~\citep{linearRAG}.
GFM-RAG trains a graph foundation model to capture query-knowledge relations over graph structure~\citep{gfmRAG}.
GraphFlow learns a retrieval policy over text-rich knowledge graphs~\citep{graphFlow}.
These methods show that graph structure is useful for complex retrieval.
LivingRAG builds on this line of work.
It adds verified experience write-back to reuse prior reasoning in later graph retrieval and generation.

\paragraph{Reasoning-Enhanced RAG.}
Another line of work improves RAG with stronger reasoning.
Iterative retrieval methods decompose a query and retrieve evidence step by step~\citep{trivedi2023interleaving,jiang2023active}.
Self-RAG uses self-reflection to decide when and how to retrieve~\citep{asai2024selfRAG}.
Chain-of-Note generates intermediate notes over retrieved evidence~\citep{chainofNote}.
LogicRAG decomposes complex questions into subqueries according to logical dependencies~\citep{logicRAG}.
These methods mainly improve reasoning for the current query.
LivingRAG instead studies how useful reasoning from earlier queries can be stored and reused in later queries.

\paragraph{Memory-augmented LLMs and RAG.}
Memory-based language agents store past observations, feedback, or plans to improve later decisions~\citep{reflexion,selfRefine}.
Recent systems also study memory for retrieval and reasoning.
REMem builds a hybrid episodic memory graph from interaction histories and uses an agentic retriever for episodic recollection and reasoning~\citep{shu2026remem}.
MemoRAG builds a global memory for long-context processing and uses it to generate retrieval clues~\citep{MemoRAG}.
LivingRAG is complementary to these methods.
It targets online Graph RAG over QA streams.
Its stored unit is a verified reasoning experience.
Each experience contains a sparse activation map, a compact reasoning summary, and validation metadata.
The activation map guides later graph retrieval.
The summary provides a reasoning scaffold for generation.
A candidate is written only when it passes grounding and novelty checks.

\section{Experimental Details}
\label{app:experimental-details}

\paragraph{Entity extraction model.}
For experiments that use spaCy-based entity extraction, we use the English transformer pipeline \texttt{en\_core\_web\_trf}.

\paragraph{NLI grounding model.}
For the grounding check in experience write-back, we use the \texttt{cross-encoder/nli-deberta-v3-large}. The model is used as the NLI cross-encoder to score entailment between retrieved evidence and extracted claims.

\paragraph{Parameter setting.}
For fair comparison, LinearRAG and LivingRAG use the same LinearRAG retrieval settings. LivingRAG keeps the original LinearRAG parameters unchanged and only adds the experience-store components described in Section~\ref{sec:method}.

\section{Detailed Complexity Analysis}
\label{app:complexity-analysis}

This appendix summarizes the extra cost introduced by LivingRAG over LinearRAG.
Let $N_t=|\mathcal{E}_t|$ be the number of stored experiences before query $q_t$, $d$ the embedding dimension, $b_t$ the number of nonzero entities in the current base activation vector, and $\bar{z}$ the average number of nonzero entries in a stored activation map.
LivingRAG does not rebuild or modify the LinearRAG corpus graph.
It only changes the initial entity activation vector and then calls the same graph retrieval procedure.
Thus, the corpus-side construction and retrieval complexity remains the LinearRAG complexity.

The main added online cost is the exact scan over the experience store.
For each stored experience, LivingRAG computes one dense query-embedding similarity and one sparse activation-map similarity:
\begin{equation}
O\bigl(N_t(d+b_t+\bar{z})\bigr)
\approx
O\bigl(N_t(d+\bar{z})\bigr).
\end{equation}
After matching, only the top-$K$ experiences are fused into the seed activation vector.
If each selected activation map is optionally truncated to its top $H$ entities, the fusion cost is
\begin{equation}
O\bigl(K\min(\bar{z},H)\bigr),
\end{equation}
where $K$ and $H$ are fixed hyperparameters.
Template-aware scaffold selection, when enabled, adds at most one additional store scan.
Template embeddings may be computed lazily from stored scaffold or source question text and cached transiently, but they are not part of persistent experience storage.

Write-back validation is also store-side and selective.
Novelty Score (NS) uses the same dual-channel scan, $O(N_t(d+\bar{z}))$.
Grounding Score (GS) is computed only after NS passes.
If $c_t$ claims are extracted and the evidence pre-filter keeps $K_g$ candidate evidence sentences per claim, the NLI stage costs
\begin{equation}
O(c_tK_g),
\end{equation}
where $K_g$ is fixed and the evidence comes only from the small set of retrieved passages.
The combined per-query cost can therefore be written as
\begin{equation}
\begin{aligned}
T_{\mathrm{Living}}(q_t)
= {}&
T_{\mathrm{Linear}}(q_t)\\
&+
O\bigl(N_t(d+\bar{z})\bigr)\\
&+
O\bigl(K\min(\bar{z},H)\bigr)\\
&+
\mathbb{1}_{\mathrm{GS}}O(c_tK_g).
\end{aligned}
\end{equation}
The expression separates the unchanged corpus-side LinearRAG cost from the added experience-store cost.

Over a stream of $Q$ queries, exact store scanning contributes $\sum_{t=1}^{Q}N_t$.
The worst case is quadratic in $Q$ if every query is stored, but LivingRAG does not store every candidate.
Across our five streams, only 1806 of 5817 candidates pass NS and reach GS.
The other 4011 candidates skip the NLI check.
Finally, 1592 candidates are written to the store, giving a final write-back ratio of 27.4\%.
Thus, store growth is tied to verified and novel experiences rather than to the raw number of queries.

The additional storage is similarly compact.
Each stored experience contains one query embedding, a sparse activation map, a short scaffold, and metadata.
Let $\bar{\ell}$ denote the average scaffold length.
The experience-store overhead is
\begin{equation}
S_{\mathrm{Living}}(t)
=
S_{\mathrm{Linear}}
+
O\bigl(N_t(d+\bar{z}+\bar{\ell})\bigr).
\end{equation}
Template embeddings, when used for scaffold selection, are transient cached values and are not counted in this persistent storage term.
Crucially, LivingRAG stores sparse activation maps rather than dense graph-entity vectors, and stores compact scaffolds rather than full reasoning traces.
The extra space is therefore linear in the number of accepted experiences and remains separate from the static corpus graph.
Overall, LivingRAG preserves LinearRAG's corpus-side scalability while adding a controlled experience-store overhead that is bounded by sparsification, top-$K$ fusion, and selective write-back.
Appendix~\ref{app:runtime-breakdown} reports the measured runtime of these components.
Appendix~\ref{app:eqs-gates} explains why grounding checks are run only for a subset of candidates.

\section{Quality Gates for Experience Write-Back}
\label{app:eqs-gates}

\begin{figure*}[t]
\centering
\includegraphics[width=\textwidth]{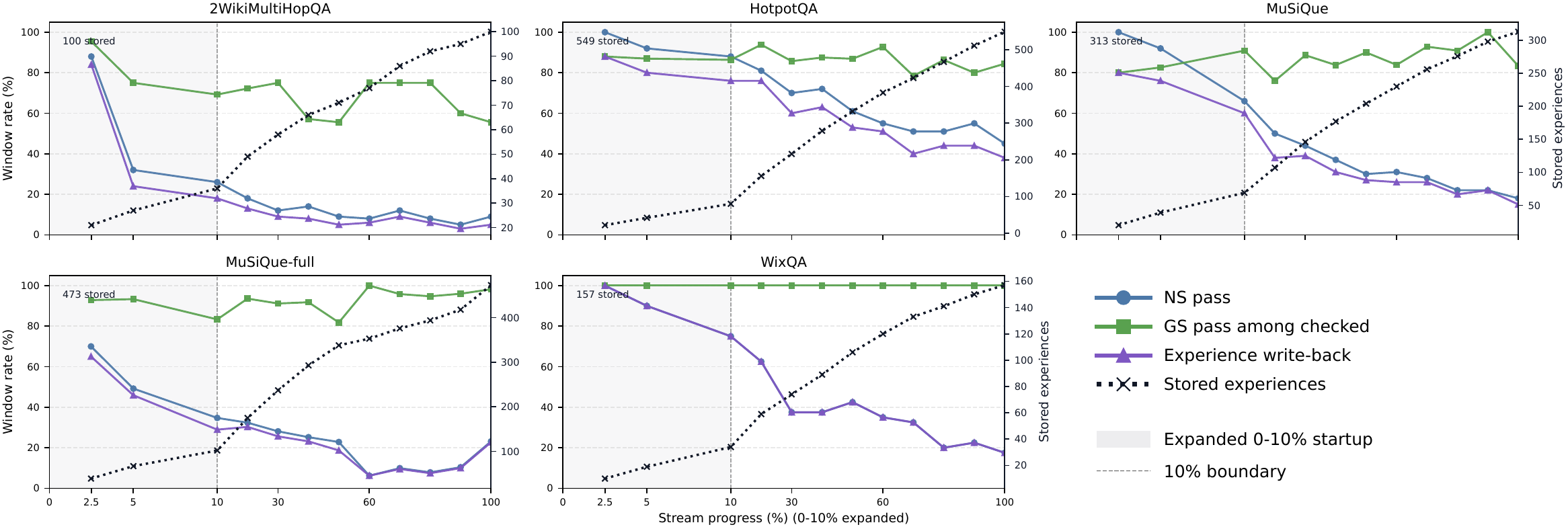}
\caption{
Dynamics of the NS and GS gates during online inference.
NS pass and experience write-back are computed over each window.
GS pass is computed only over candidates that reach the GS check.
The black curve shows the cumulative number of stored experiences.
The first 10\% of each stream is expanded to show the startup stage.
}
\label{fig:eqs_gate_dynamics}
\end{figure*}

\begin{table*}[t]
\centering
\small
\setlength{\tabcolsep}{4pt}
\caption{Effect of the NS and GS gates on experience write-back. All runs use $\tau_{\mathrm{novel}}=0.7$ and $\tau_{\mathrm{gs}}=0.6$. NS pass is measured over all questions. GS pass is measured only on candidates that reached the GS check. GS reject counts the checked candidates that failed the grounding gate.}
\label{tab:eqs_gate_summary}
\begin{tabular}{lrrrrrr}
\toprule
\textbf{Dataset} & \textbf{Questions} & \textbf{NS pass (\%)} & \textbf{GS checked} & \textbf{GS pass (\%)} & \textbf{GS reject} & \textbf{Stored experiences} \\
\midrule
2Wiki & 1000 & 13.80 & 138 & 72.46 & 38 & 100 \\
HotpotQA & 1000 & 63.30 & 633 & 86.73 & 84 & 549 \\
MuSiQue & 1000 & 36.30 & 363 & 86.23 & 50 & 313 \\
MuSiQue-full & 2417 & 21.31 & 515 & 91.84 & 42 & 473 \\
WixQA & 400 & 39.25 & 157 & 100.00 & 0 & 157 \\
\midrule
All & 5817 & 31.05 & 1806 & 88.15 & 214 & 1592 \\
\bottomrule
\end{tabular}
\end{table*}

Figure~\ref{fig:eqs_gate_dynamics} and Table~\ref{tab:eqs_gate_summary} analyze the two gates used before a candidate is written as a verified experience.
For implementation efficiency, LivingRAG evaluates NS before GS.
The change affects only the order of computation.
A new experience is stored only when both gates pass.
When NS fails, the system skips GS and avoids the NLI check.

The NS curve is high at the beginning because there are few stored experiences.
As more experiences are written, the NS pass rate decreases.
The write-back curve follows the same trend.
Across the five streams, only 1806 of 5817 candidates pass NS.
The number of GS checks is also 1806, which confirms that GS is skipped when NS fails.
The final number of stored experiences is 1592.
The count shows that LivingRAG does not store every query.
It mainly stores candidates that remain novel relative to earlier experiences.
GS is then applied to the smaller candidate set.
In total, 4011 candidates fail NS and skip GS.
Among the 1806 candidates checked by GS, 1592 pass and 214 are rejected.
The GS pass rate among checked candidates is 88.15\%.
The results show that NS controls store growth.
GS removes ungrounded candidates before write-back.

WixQA is a special case in Table~\ref{tab:eqs_gate_summary}.
Its GS pass rate is 100\% among the candidates that reach the GS check.
The value does not mean that all GS scores are saturated.
The average GS on the checked WixQA candidates is 0.886, and the lowest score is 0.653.
Thus, the result means that all checked candidates are above the grounding threshold of 0.6.
The likely reason is that the WixQA corpus consists of official help documents.
Many answers contain procedural information that is directly supported by the retrieved help documents.
For this dataset, we therefore use GS as a light grounding constraint rather than as the main selective gate.
In parameter tuning, this corresponds to keeping the grounding threshold permissive and relying on the novelty gate and WixQA-specific retrieval settings to control write-back.
The effective selectivity is shifted toward NS, which filters repeated support patterns before GS is run.
The setting prevents WixQA from being dominated by grounding rejection, while still requiring each stored experience to have explicit passage support.

\begin{table*}[t]
\centering
\scriptsize
\setlength{\tabcolsep}{2pt}
\caption{Representative NS and GS case studies. The cases are selected from the final LivingRAG runs without re-running retrieval or validation.}
\label{tab:eqs_case_study}
\begin{tabular}{@{}P{0.09\linewidth}P{0.065\linewidth}rP{0.18\linewidth}P{0.135\linewidth}rrP{0.095\linewidth}P{0.22\linewidth}@{}}
\toprule
\textbf{Case} & \textbf{Dataset} & \textbf{Idx.} & \textbf{Query} & \textbf{Answer} & \textbf{NS} & \textbf{GS} & \textbf{Decision} & \textbf{Interpretation} \\
\midrule
Stored multi-hop candidate & HotpotQA & 33 & What subsidiary of the largest airline of the Republic of China (Taiwan) has a main hub at Taichung Airport? & Mandarin Airlines & 0.793 & 1.000 & Stored & Both gates pass. The retrieved airline passage directly supports the answer, so the candidate becomes a verified experience. \\
Direct WixQA grounding & WixQA & 57 & I would like to know how many sites I can have. & There is no limit to the number of sites you can create under a single Wix account; you can have as many as you want. & 0.829 & 1.000 & Stored & The answer contains directly supported procedural information from the retrieved help articles. This illustrates why WixQA GS scores are often high under the permissive grounding threshold. \\
Novelty rejection & WixQA & 391 & can I get a refund on the domain & No. Wix domain purchases are non-refundable and do not include a 14-day money-back guarantee or trial period. If you no longer... & 0.213 & -- & Rejected by NS & The query repeats a stored domain-refund experience. NS assigns a low score, skips GS, and avoids storing another copy of the same support pattern. \\
Grounding rejection & HotpotQA & 858 & Spawn, a fictional character who Mustafa Shakir has expressed interest in playing, was created by? & Todd McFarlane & 0.701 & 0.368 & Rejected by GS & NS passes, but the retrieved passages do not mention the key entities. The answer matches the gold label, yet GS blocks write-back because the candidate is not grounded in retrieved evidence. \\
\bottomrule
\end{tabular}
\end{table*}

Table~\ref{tab:eqs_case_study} gives concrete examples of these decisions.
The two stored cases show when a candidate is written to the store.
It must be new enough and supported by retrieved passages.
The WixQA grounding case also explains why this dataset can have high GS pass rates.
Its answer is stated directly in the help articles.
The NS rejection case shows the opposite behavior.
A later domain-refund query retrieves an almost identical stored experience.
NS assigns a very low score and prevents another copy from being written.
Finally, the GS rejection case shows why grounding is still needed after NS.
The model predicts the correct answer, but the retrieved passages do not contain the key entities needed to support it.
LivingRAG therefore returns the answer but refuses to store the candidate.
The grounding gate prevents unsupported reasoning from becoming reusable experience.
The selectivity also explains the validation term in the complexity analysis in Appendix~\ref{app:complexity-analysis}.

\section{Dataset Details}
\label{app:dataset-details}

We evaluate LivingRAG on five online QA streams.
Three of them are standard multi-hop QA benchmarks: 2WikiMultiHopQA, HotpotQA, and MuSiQue.
They mainly test whether a system can retrieve and combine evidence across multiple facts.
Most answers in these datasets are short entities, dates, locations, or yes/no labels.

We further include MuSiQue-full and WixQA.
These two datasets test different aspects of online experience reuse.
MuSiQue-full provides a larger multi-hop stream with denser reuse structure.
WixQA provides a customer-support stream over a shared enterprise knowledge base.
Together, they make the evaluation less dependent on a single dataset style.

\paragraph{MuSiQue-full.}
MuSiQue is built by composing single-hop questions into multi-hop questions.
It provides decomposition annotations that expose which single-hop questions are reused across different multi-hop questions.
This makes it useful for studying whether an experience-reusing RAG system can reuse prior reasoning steps.

Prior RAG evaluations often use a 1,000-question MuSiQue subset.
We additionally evaluate on a larger answerable split, which we call MuSiQue-full.
It contains 2,417 answerable questions.
This name is used only to distinguish it from the smaller subset in our experiments.
It does not refer to the official unanswerable MuSiQue-Full setting.

We define an experience cluster using the official MuSiQue decomposition.
Two questions are linked if they share at least one decomposed single-hop question.
We then compute connected components over this question graph.
A component with at least two questions is treated as an experience cluster.
This definition is based on the dataset structure.
It does not use text similarity, embedding similarity, or model predictions.

\begin{table*}[t]
\centering
\small
\setlength{\tabcolsep}{3.8pt}
\caption{
Reuse structure in the MuSiQue streams.
Experience clusters are connected components of questions that share at least one decomposed single-hop question.
}
\label{tab:musique_full_stats}
\begin{tabular}{lrrrrrr}
\toprule
\textbf{Dataset}
& \textbf{\#Q}
& \textbf{\#Chunks}
& \textbf{\#Clusters}
& \textbf{Coverage}
& \textbf{Avg. Size}
& \textbf{Max Size} \\
\midrule
MuSiQue      & 1,000 & 1,354 & 178 & 79.2 & 4.4 & 21 \\
MuSiQue-full & 2,417 & 2,379 & 300 & 89.9 & 7.2 & 58 \\
\bottomrule
\end{tabular}
\end{table*}

Table~\ref{tab:musique_full_stats} shows that MuSiQue-full contains a stronger reuse structure than the smaller MuSiQue subset.
It has more clusters, higher cluster coverage, and larger average cluster size.
This makes it a useful stress test for online experience reuse.
A system that stores verified experiences should benefit more when later questions reuse earlier reasoning components.

\paragraph{WixQA.}
WixQA is different from the Wikipedia-style benchmarks.
It is built from product-support questions over the Wix Help Center.
The questions are about product features, account settings, payments, domains, site editing, email, bookings, and related user tasks.
The answers are usually support-style solutions.
They often describe what the user should do, which product should be used, or whether a feature is supported.

This setting is important for LivingRAG.
All questions are grounded in the same enterprise knowledge base.
Many user issues refer to related products, settings, and help articles.
This creates repeated evidence neighborhoods and recurring support patterns.
At the same time, the language is less formulaic than standard multi-hop QA.
This makes WixQA a useful test for experience reuse in a more deployment-like setting.

We use 400 WixQA questions in the main evaluation.
The sample is balanced between ExpertWritten and Simulated questions.
We exclude the Synthetic subset because it is generated and more tightly tied to individual articles.
The retrieval corpus is the Wix Help Center corpus.
Following the LinearRAG setting, articles are segmented into chunks of roughly 600 tokens.

\section{WixQA-specific Processing and Evaluation}
\label{app:wixqa-processing}

WixQA requires several dataset-specific choices.
These choices are not changes to the LivingRAG framework.
They are needed because WixQA differs from the other datasets in entity type, passage length distribution, and answer format.
In the experiments that compare LinearRAG and LivingRAG on WixQA, both systems use the same WixQA-specific processing.

\subsection{LLM-based Entity Extraction}
\label{app:wixqa-llm-ner}

The graph retriever relies on query entities to initialize graph activation.
For Wikipedia-style QA, a general-purpose NER model is usually sufficient.
Most questions contain people, locations, organizations, works, dates, or named events.
These entities are also the main anchors of the reasoning chain.
For example, in 2WikiMultiHopQA and HotpotQA, spaCy often extracts useful entities such as person names, film titles, places, and organizations.

WixQA has a different language style.
Its important query cues are often product names, UI labels, plan names, feature names, settings, and user actions.
These cues are not ordinary named entities.
They are domain terms from the Wix Help Center.
If they are missed, graph retrieval starts from weak or empty query seeds.

Table~\ref{tab:wixqa_ner_examples} compares representative cases.
For standard multi-hop QA, spaCy usually extracts useful named entities.
For WixQA, spaCy often returns no entity or only a broad brand-level entity.
LLM NER instead extracts product-domain terms that are more useful for support retrieval.

\begin{table*}[t]
\centering
\small
\setlength{\tabcolsep}{3.0pt}
\caption{
Examples of query-side entity extraction.
spaCy works well for standard multi-hop QA because the key retrieval cues are ordinary named entities.
On WixQA, the key cues are product-domain terms, so spaCy often fails or extracts only broad entities.
}
\label{tab:wixqa_ner_examples}
\begin{tabular}{p{0.12\linewidth}p{0.35\linewidth}p{0.18\linewidth}p{0.25\linewidth}}
\toprule
\textbf{Dataset}
& \textbf{Question}
& \textbf{spaCy NER}
& \textbf{LLM NER or useful seeds} \\
\midrule

2Wiki
& When did Lothair II's mother die?
& \textit{Lothair II}
& Not needed \\

\addlinespace

2Wiki
& Which film was released first, Aas Ka Panchhi or Phoolwari?
& \textit{Aas Ka Panchhi}; \textit{Phoolwari}
& Not needed \\

\addlinespace

HotpotQA
& In which county is the town in which Raymond Robertsen was born?
& \textit{Raymond Robertsen}
& Not needed \\

\addlinespace

MuSiQue
& When was the person who Messi's goals in Copa del Rey compared to get signed by Barcelona?
& \textit{Messi}; \textit{Copa del Rey}; \textit{Barcelona}
& Not needed \\

\midrule

WixQA
& I'm want to know how much it would cost to upgrade my email plan.
& \texttt{[]}
& \textit{Email Marketing Plan}; \textit{upgrade Email Marketing Plan}; \textit{Email Marketing Plan pricing} \\

\addlinespace

WixQA
& How do I sync the hotel app with my calendars ical link to allow visitors to book available dates on my site?
& \texttt{[]}
& \textit{Wix Bookings}; \textit{iCal URL}; \textit{syncing calendars}; \textit{booking available dates} \\

\addlinespace

WixQA
& I have not received my payment in my bank account yet.
& \texttt{[]}
& \textit{Wix Payments}; \textit{payout delay}; \textit{failed payout}; \textit{Payouts on Hold} \\

\addlinespace

WixQA
& I want to know if the Wix store function work for selling services instead of just physical goods.
& \textit{Wix}
& \textit{Wix Stores}; \textit{selling services}; \textit{service bookings}; \textit{Wix Bookings} \\

\bottomrule
\end{tabular}
\end{table*}

These examples show that the issue is not only low recall.
It is also a mismatch between general NER labels and product-support terminology.
For example, the last WixQA question cannot be answered by recognizing only the word \textit{Wix}.
The retrieval system must distinguish whether the user should use \textit{Wix Stores} or \textit{Wix Bookings}.
This distinction is central to the answer.

We therefore use LLM-based entity extraction for WixQA.
The extractor is implemented with Qwen3.5-9B.
This is a relatively small LLM, but we find it sufficient for this task in practice.
The task only requires extracting short product-domain seed terms, not solving the question.
Thus, a smaller LLM can provide useful domain entities without changing the retrieval framework.
For fairness, LinearRAG and LivingRAG use the same passage-level and query-level extractor in all WixQA runs.

The goal is not to add reasoning to the retriever.
The goal is to produce better product-domain seed terms.
The downstream retrieval pipeline remains unchanged.
Extracted query terms are mapped to graph entities and then used by the same graph propagation and passage ranking steps.

\begin{figure*}[t]
\centering
\begin{promptbox}{Passage-level LLM NER prompt}
System:
You are an entity extraction assistant for Wix product documentation. Extract ALL of the following from the given text:

1. Named product / feature / UI names (e.g. 'Wix Events', 'Event Details Page', 'Dashboard', 'DNS records', 'CMS collection')
2. The main topic or action concept of this article as 1-3 short phrases that would match a user's question about this content (e.g. 'Wix Payments account verification', 'adding funds to cover refund', 'publishing pages to live site', 'staff working hours', 'CMS collection item order')

Return ONLY a JSON array combining all extracted terms. No explanation, no markdown fences.

Example -- article about verifying a Wix Payments account:
["Wix Payments", "Wix Payments account verification", "account verification status", "Accept Payments", "Adyen", "Stripe"]

User:
{passage}

Response:
\end{promptbox}
\caption{
Passage-level prompt used for WixQA LLM-based entity extraction during indexing.
The extractor identifies Wix product-documentation entities and article-level action-topic terms from each passage.
}
\label{fig:wixqa_passage_llm_ner_prompt}
\end{figure*}

\begin{figure*}[t]
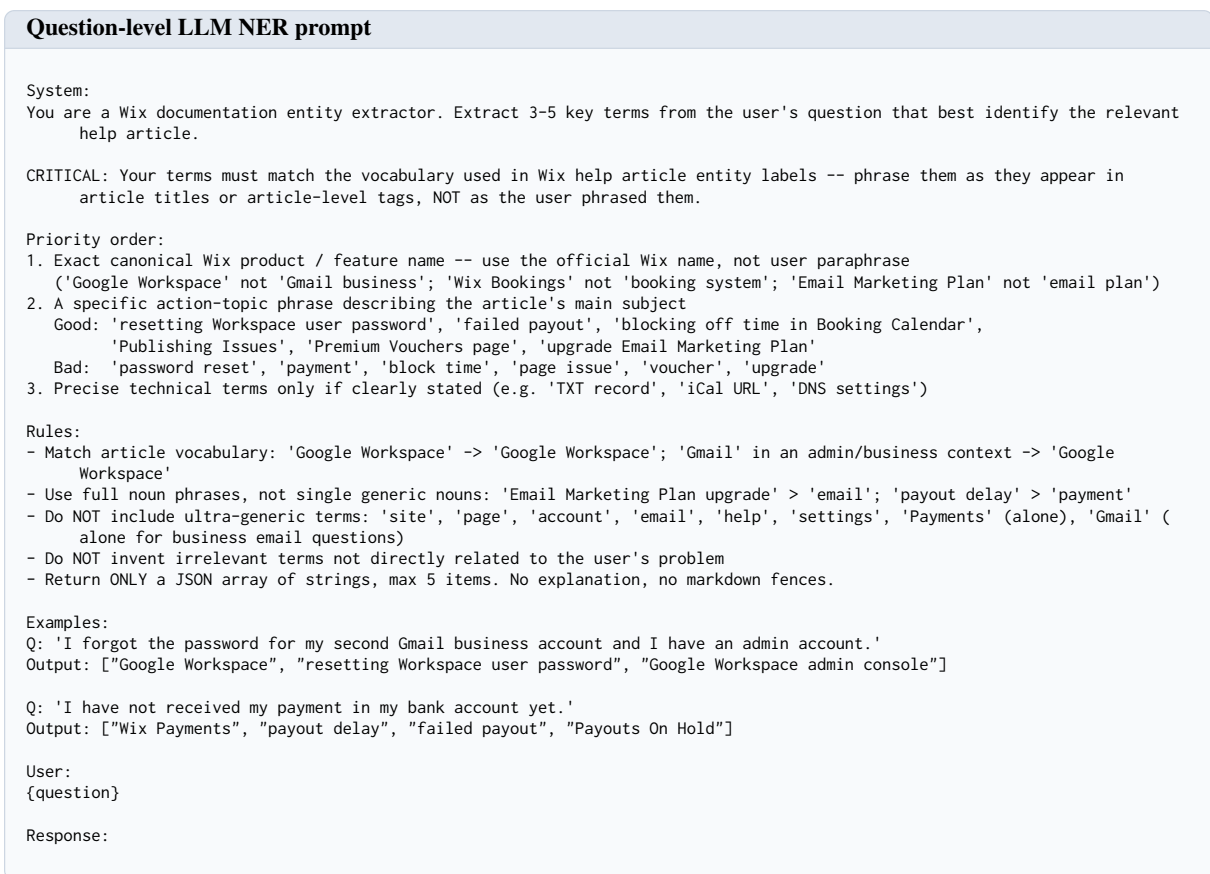

\centering
\begin{promptbox}{Question-level LLM NER prompt}
System:
You are a Wix documentation entity extractor. Extract 3-5 key terms from the user's question that best identify the relevant help article.

CRITICAL: Your terms must match the vocabulary used in Wix help article entity labels -- phrase them as they appear in article titles or article-level tags, NOT as the user phrased them.

Priority order:
1. Exact canonical Wix product / feature name -- use the official Wix name, not user paraphrase
   ('Google Workspace' not 'Gmail business'; 'Wix Bookings' not 'booking system'; 'Email Marketing Plan' not 'email plan')
2. A specific action-topic phrase describing the article's main subject
   Good: 'resetting Workspace user password', 'failed payout', 'blocking off time in Booking Calendar',
         'Publishing Issues', 'Premium Vouchers page', 'upgrade Email Marketing Plan'
   Bad:  'password reset', 'payment', 'block time', 'page issue', 'voucher', 'upgrade'
3. Precise technical terms only if clearly stated (e.g. 'TXT record', 'iCal URL', 'DNS settings')

Rules:
- Match article vocabulary: 'Google Workspace' -> 'Google Workspace'; 'Gmail' in an admin/business context -> 'Google Workspace'
- Use full noun phrases, not single generic nouns: 'Email Marketing Plan upgrade' > 'email'; 'payout delay' > 'payment'
- Do NOT include ultra-generic terms: 'site', 'page', 'account', 'email', 'help', 'settings', 'Payments' (alone), 'Gmail' (alone for business email questions)
- Do NOT invent irrelevant terms not directly related to the user's problem
- Return ONLY a JSON array of strings, max 5 items. No explanation, no markdown fences.

Examples:
Q: 'I forgot the password for my second Gmail business account and I have an admin account.'
Output: ["Google Workspace", "resetting Workspace user password", "Google Workspace admin console"]

Q: 'I have not received my payment in my bank account yet.'
Output: ["Wix Payments", "payout delay", "failed payout", "Payouts On Hold"]

User:
{question}

Response:
\end{promptbox}
\caption{
Question-level prompt used for WixQA LLM-based entity extraction.
The extractor maps user questions to canonical Wix product names and action-topic phrases for query-side graph activation.
}
\label{fig:wixqa_question_llm_ner_prompt}
\end{figure*}

Figure~\ref{fig:wixqa_passage_llm_ner_prompt} shows the passage-level prompt used during indexing.
Figure~\ref{fig:wixqa_question_llm_ner_prompt} shows the question-level prompt used for query-side entity extraction.

\begin{table*}[t]
\centering
\small
\setlength{\tabcolsep}{4.5pt}
\caption{
Query-side entity extraction statistics.
LLM NER addresses the low-recall problem of standard NER on WixQA while keeping the number of query seeds small.
}
\label{tab:wixqa_query_ner}
\begin{tabular}{llrrrr}
\toprule
\textbf{Method}
& \textbf{Query Set}
& \textbf{0 Ent.}
& \textbf{Avg.}
& \textbf{Med.}
& \textbf{Range} \\
\midrule
spaCy NER
& 2Wiki, first 400
& 1.2
& 1.68
& 2
& 0--4 \\
spaCy NER
& WixQA, all 400
& 52.5
& 0.61
& 0
& 0--4 \\
LLM NER + mapping
& WixQA, all 400
& 0.0
& 3.77
& 4
& 1--5 \\
\bottomrule
\end{tabular}
\end{table*}

Table~\ref{tab:wixqa_query_ner} shows that the examples in Table~\ref{tab:wixqa_ner_examples} are not isolated cases.
On 2WikiMultiHopQA, spaCy almost always extracts at least one query entity.
On WixQA, it extracts no entity for 52.5\% of the questions.
This weakens graph retrieval because many queries have no useful starting nodes.
LLM NER removes this failure case.

LLM NER is not noise-free.
Some extracted terms can be mapped to imperfect graph entities.
We therefore only keep the top mapped candidates for each query term and remove high-degree hub entities.
This keeps query activation focused while still giving the retriever product-specific seeds.

We also apply the LLM extractor to WixQA passages during indexing.
This keeps the entity space consistent between queries and passages.
The resulting graph remains sparse enough for propagation.
Almost every chunk has extracted entities, and the average number of entities per entity-bearing sentence is below two.

\begin{table}[t]
\centering
\small
\setlength{\tabcolsep}{5pt}
\caption{
Index-side entity extraction statistics for WixQA with LLM NER.
}
\label{tab:wixqa_index_ner}
\begin{tabular*}{\columnwidth}{@{\extracolsep{\fill}}lr@{}}
\toprule
\textbf{Statistic} & \textbf{Value} \\
\midrule
Chunks & 7,132 \\
Chunks with no entity & 1 \\
Avg. entities per chunk & 11.23 \\
Median entities per chunk & 10 \\
Sentences with entities & 68,798 \\
Avg. entities per entity-bearing sentence & 1.79 \\
Median entities per entity-bearing sentence & 1 \\
\bottomrule
\end{tabular*}
\end{table}

\subsection{Passage Length Normalization}

WixQA also differs in passage length distribution.
The Wikipedia-style datasets are mostly segmented into near-fixed long chunks.
In contrast, WixQA keeps the natural granularity of help-center articles.
Some chunks are short feature notes or known-issue pages.
Others are long tutorials with detailed steps.

This creates a different retrieval problem.
Long chunks naturally contain more words and more entity mentions.
They can therefore receive higher retrieval scores because they cover more topics.
Short chunks may be more precise, but they provide fewer matching signals.
Without correction, a long generic article may dominate a short article that directly answers the user issue.

\begin{table}[t]
\centering
\small
\setlength{\tabcolsep}{2.5pt}
\caption{
Passage length distribution by character count.
WixQA contains many short chunks, while the Wikipedia-style datasets are close to fixed-length chunks.
}
\label{tab:wixqa_length_distribution}
\begin{tabular*}{\columnwidth}{@{\extracolsep{\fill}}lrrrr@{}}
\toprule
\textbf{Dataset}
& \textbf{\#Chk.}
& \textbf{Median}
& \textbf{$<1000$}
& \textbf{$<2000$} \\
\midrule
WixQA & 7,132 & 1,321 & 44.4 & 59.6 \\
2Wiki & 658 & 4,449 & 0.0 & 0.0 \\
HotpotQA & 1,311 & 4,651 & 0.0 & 0.0 \\
MuSiQue & 1,354 & 4,701 & 0.0 & 0.0 \\
\bottomrule
\end{tabular*}
\end{table}

Table~\ref{tab:wixqa_length_distribution} shows this difference.
In WixQA, 44.4\% of chunks contain fewer than 1,000 characters, and 59.6\% contain fewer than 2,000 characters.
The corresponding ratios are almost zero in the other datasets.
This means that WixQA retrieval must handle a mix of very short and long passages.

We apply a weak BM25-style length normalization to WixQA passage scores:
\begin{equation}
\mathrm{score}'(p)
=
\frac{\mathrm{score}(p)}
{(1-b)+b\cdot |p|/\overline{|p|}} .
\end{equation}
Here $|p|$ is the passage length, and $\overline{|p|}$ is the average passage length in the corpus.
We set $b=0.1$.
This value is intentionally small.
It mildly corrects length bias, but it does not strongly favor short passages.
This is important because some long Wix articles contain necessary step-by-step evidence.

\subsection{Evaluation Metric for WixQA}
\label{app:wixqa-eval}

We do not use contain-match accuracy as the main metric for WixQA.
Contain-match accuracy is suitable for short-answer QA.
In 2WikiMultiHopQA, HotpotQA, and MuSiQue, the gold answer is usually a short entity, date, place, or yes/no label.
A correct generated answer often contains the exact gold answer span.

WixQA is different.
Its gold answers are support-style solutions.
They often include conditions, product-specific guidance, operation steps, caveats, and official wording.
A generated answer may solve the user's problem without copying the full gold answer.
Thus, requiring the full gold answer to appear as a substring is too strict.

Table~\ref{tab:answer_format_examples} gives representative examples.
The standard multi-hop QA answers are short spans.
In contrast, WixQA answers describe how to solve a user-support problem.
The answer may be 60, 120, or more than 300 words.

\begin{table*}[t]
\centering
\footnotesize
\setlength{\tabcolsep}{2pt}
\caption{
Gold answer format differs sharply between standard multi-hop QA and WixQA.
Standard QA usually expects a short span.
WixQA expects a support solution.
}
\label{tab:answer_format_examples}
\begin{tabular}{@{}P{0.14\linewidth}P{0.44\linewidth}P{0.27\linewidth}r@{}}
\toprule
\textbf{Dataset}
& \textbf{Question}
& \textbf{Gold answer}
& \textbf{Words} \\
\midrule

2Wiki
& When did Lothair II's mother die?
& 20 March 851
& 3 \\

\addlinespace

2Wiki
& Which film was released first, Aas Ka Panchhi or Phoolwari?
& Phoolwari
& 1 \\

\addlinespace

HotpotQA
& In which county is the town in which Raymond Robertsen was born?
& Finnmark county
& 2 \\

\addlinespace

MuSiQue
& When was the person who Messi's goals in Copa del Rey compared to get signed by Barcelona?
& June 1982
& 2 \\

\midrule

WixQA
& Can I start accepting payments on my site while my Wix Payments account is still under verification?
& You can start accepting payments almost immediately, but Wix must verify your identity before the account is fully activated.
& 27 \\

\addlinespace

WixQA
& I want to know if the Wix store function work for selling services instead of just physical goods.
& The answer distinguishes \textbf{Wix Stores} from \textbf{Wix Bookings}. For selling services, it recommends Wix Bookings because it supports service booking and online payment.
& 66 \\

\addlinespace

WixQA
& I am inquiring about purchasing the yearly premium plan with a free domain for 1 year. The voucher does not show up at checkout. Do I need to purchase the plan first?
& The answer explains that the voucher is not shown at checkout. It becomes available after purchasing a yearly Premium plan and can be claimed from the Premium Vouchers page.
& 120 \\

\addlinespace

WixQA
& How can I add discounts to my service prices when customers pay for a plan?
& The answer explains how to create a Pricing Plans coupon, including discount type, applicable plans, billing-cycle behavior, dates, usage limits, and coupon creation.
& 328 \\

\bottomrule
\end{tabular}
\end{table*}

\begin{table*}[t]
\centering
\small
\setlength{\tabcolsep}{4.5pt}
\caption{
Answer length statistics.
WixQA answers are long support solutions rather than short answer spans.
}
\label{tab:wixqa_answer_length}
\begin{tabular}{lrrr}
\toprule
\textbf{Dataset}
& \textbf{Median words}
& \textbf{$\leq$5 words}
& \textbf{$>$30 words} \\
\midrule
2WikiMultiHopQA & 2 & 95.5 & 0.0 \\
HotpotQA & 2 & 96.1 & 0.0 \\
MuSiQue & 2 & 91.2 & 0.0 \\
WixQA & 59 & 0.2 & 85.0 \\
\bottomrule
\end{tabular}
\end{table*}

Table~\ref{tab:wixqa_answer_length} shows the scale of this mismatch.
The median answer length in WixQA is 59 words.
For the other datasets, the median answer length is only 2 words.
For WixQA, 85.0\% of answers are longer than 30 words.
For the other datasets, this ratio is zero.

The following example illustrates why contain-match is misleading.
A generated answer can be useful and correct even if it does not contain the full official support answer.

\begin{center}
\begin{papertextbox}[width=0.94\linewidth]{Why contain-match fails}
\small
\textbf{Question.}
How can I add discounts to my service prices when customers pay for a plan?

\vspace{1.5mm}
\textbf{Gold answer type.}
A long official support answer.
It explains how to create a Pricing Plans coupon.
It also covers discount type, applicable plans, billing-cycle behavior, validity dates, usage limits, and final coupon creation.

\vspace{1.5mm}
\textbf{Possible correct model answer.}
Go to Pricing Plans and create a coupon.
Choose a fixed or percentage discount.
Select the plan it applies to.
Then configure billing behavior, validity dates, and usage limits.

\vspace{1.5mm}
\textbf{Why contain-match fails.}
The model answer gives the correct operation.
However, it will not contain the full official answer as a substring.
\end{papertextbox}
\end{center}

This is why contain-match accuracy severely underestimates answer quality on WixQA.
In our 400-question WixQA evaluation, LinearRAG has 0 contain-match hits.
LivingRAG has only 1 contain-match hit.
However, LLM-based evaluation gives 66.00\% accuracy for LinearRAG and 70.25\% for LivingRAG.

We use LLM accuracy for all datasets, but the evaluation prompt differs by dataset type.
For standard multi-hop QA, the evaluator checks whether the generated answer contains the key gold information and does not contradict it.
For WixQA, the evaluator checks whether the answer correctly resolves the support issue.
This requires a task-oriented prompt.
The answer must recommend the right Wix product or feature.
It must describe the core operation correctly.
It must not claim that a supported feature is unsupported, or the reverse.
It does not need to reproduce the full official wording.

This difference is important in product-support QA.
For example, if a user asks whether Wix Stores can be used to sell services, the key issue is not whether the answer mentions the words \textit{Wix Stores}.
The key issue is whether it recommends the right product.
An answer that tells the user to use Wix Stores for service booking would be misleading.
A correct answer should identify \textit{Wix Bookings} as the suitable product for booking and paying for services.

\begin{figure}[t]
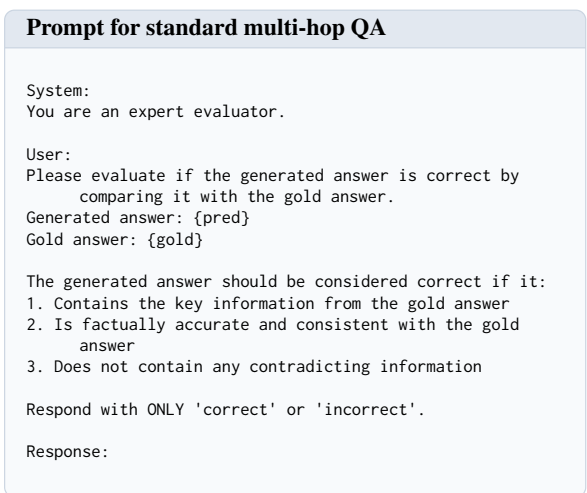

\centering
\begin{promptbox}{Prompt for standard multi-hop QA}
System:
You are an expert evaluator.

User:
Please evaluate if the generated answer is correct by comparing it with the gold answer.
Generated answer: {pred}
Gold answer: {gold}

The generated answer should be considered correct if it:
1. Contains the key information from the gold answer
2. Is factually accurate and consistent with the gold answer
3. Does not contain any contradicting information

Respond with ONLY 'correct' or 'incorrect'.

Response:
\end{promptbox}
\caption{
LLM-based accuracy evaluation prompt used for standard multi-hop QA datasets.
}
\label{fig:standard_llm_acc_prompt}
\end{figure}

\begin{figure*}[t]
\centering
\begin{promptbox}{Prompt for WixQA}
System:
You are an expert evaluator for Wix customer support answers. Your task is to judge whether a generated answer correctly resolves the customer's issue, compared to an official gold answer extracted from a Wix help article.

User:
Evaluate whether the generated answer is correct for this Wix customer support question.

Generated answer:
{pred}

Gold answer (from official Wix help article):
{gold}

Mark as CORRECT only if ALL of the following hold:
1. The same Wix product / feature is recommended (e.g. "Wix Bookings" vs "Wix Stores" is a meaningful difference).
2. The core action described would guide the customer to actually solve their problem.
3. It does not claim something is impossible when the gold shows it is possible (or vice versa).
4. It does not contain factually incorrect steps that would mislead the customer.

Mark as INCORRECT if ANY of the following hold:
- It recommends the wrong Wix product or feature.
- It describes the wrong operation (e.g. "delete a collection" vs "delete items from a collection").
- It contradicts the gold on a key fact (e.g. a feature exists vs does not exist).
- It provides outdated or clearly wrong information.

Leniency: the generated answer does NOT need to reproduce every step from the gold. Partial coverage of the correct solution is acceptable as long as the main direction is right and no incorrect information is given.

Respond with ONLY 'correct' or 'incorrect'.
\end{promptbox}
\caption{
LLM-based accuracy evaluation prompt used for WixQA.
The prompt evaluates whether a generated support answer resolves the user issue without recommending the wrong Wix product, operation, or unsupported behavior.
}
\label{fig:wixqa_llm_acc_prompt}
\end{figure*}

\section{Runtime Breakdown}
\label{app:runtime-breakdown}

We further measure the runtime of the main online components.
We use the same concurrent setting as the main experiments.
The system answers up to 16 questions in parallel.
We exclude one-time indexing, model loading, and evaluation-only LLM calls.
The reported values are per-question averages.
They are computed by summing the measured time of each component and dividing it by the number of questions.
Because the system runs concurrently, these values are module-level averages.
They should not be read as serialized wall-clock latency.
The corresponding asymptotic cost is analyzed in Appendix~\ref{app:complexity-analysis}.
Gate selectivity is analyzed in Appendix~\ref{app:eqs-gates}.

The implementation keeps the LinearRAG retrieval path unchanged.
For each query, entity matching and graph retrieval are run on CPU in the same way as LinearRAG.
LLM calls are issued concurrently.
For LivingRAG, answer generation and experience write-back are separated.
After an answer is generated, the system forms a candidate experience.
The candidate is then checked by the novelty and grounding gates described in Section~\ref{sec:experience-write-back}.
Grounding uses batched NLI cross-encoder calls.
Accepted candidates are written to the store in order.
The asynchronous design keeps online answering separate from experience-store maintenance.

\begin{table*}[t]
\centering
\small
\setlength{\tabcolsep}{6pt}
\caption{LLM, retrieval, and validation times are module-level averages.
Indexing, model loading, and evaluation-only LLM calls are excluded.
The Retrieval / LLM column reports the ratio between retrieval time and LLM time.
Validation refers to the post-answer novelty and grounding checks used for experience write-back.}
\label{tab:runtime_breakdown}
\begin{tabular}{llrrrr}
\toprule
\textbf{Dataset} & \textbf{Method} & \textbf{LLM s/q} & \textbf{Retrieval s/q} & \textbf{Retrieval / LLM} & \textbf{Validation s/q} \\
\midrule
2Wiki & LinearRAG & 30.07 & 0.50 & 1.7\% & N/A \\
2Wiki & LivingRAG & 24.04 & 0.54 & 2.2\% & 2.13 \\
HotpotQA & LinearRAG & 15.42 & 0.72 & 4.7\% & N/A \\
HotpotQA & LivingRAG & 14.45 & 0.74 & 5.1\% & 2.35 \\
MuSiQue & LinearRAG & 37.23 & 1.49 & 4.0\% & N/A \\
MuSiQue & LivingRAG & 33.88 & 1.50 & 4.4\% & 6.64 \\
\bottomrule
\end{tabular}
\end{table*}

Table~\ref{tab:runtime_breakdown} shows that retrieval is much cheaper than LLM inference.
Across the measured datasets, retrieval takes about 1.7\% to 5.1\% of the LLM time.
The retrieval time of LivingRAG is close to that of LinearRAG.
The result supports the complexity analysis.
The added experience-augmented retrieval step is not the main runtime bottleneck.

Write-back validation is more expensive than retrieval.
The cost comes from local NLI validation and experience write-back.
All NLI checks are run on a single NVIDIA GeForce RTX 3060 GPU.
The RTX 3060 is a modest GPU for cross-encoder validation.
More importantly, write-back validation is a post-answer maintenance step.
The answer has already been generated before this step starts.
In a serving system, the answer can be returned before this experience update finishes.
Thus, the validation column describes the cost of maintaining a reliable experience store.
It does not need to increase user-visible answer latency.

\section{Prompt Token Trends}
\label{app:prompt-token-trends}

\begin{figure*}[t]
\centering
\includegraphics[width=\textwidth]{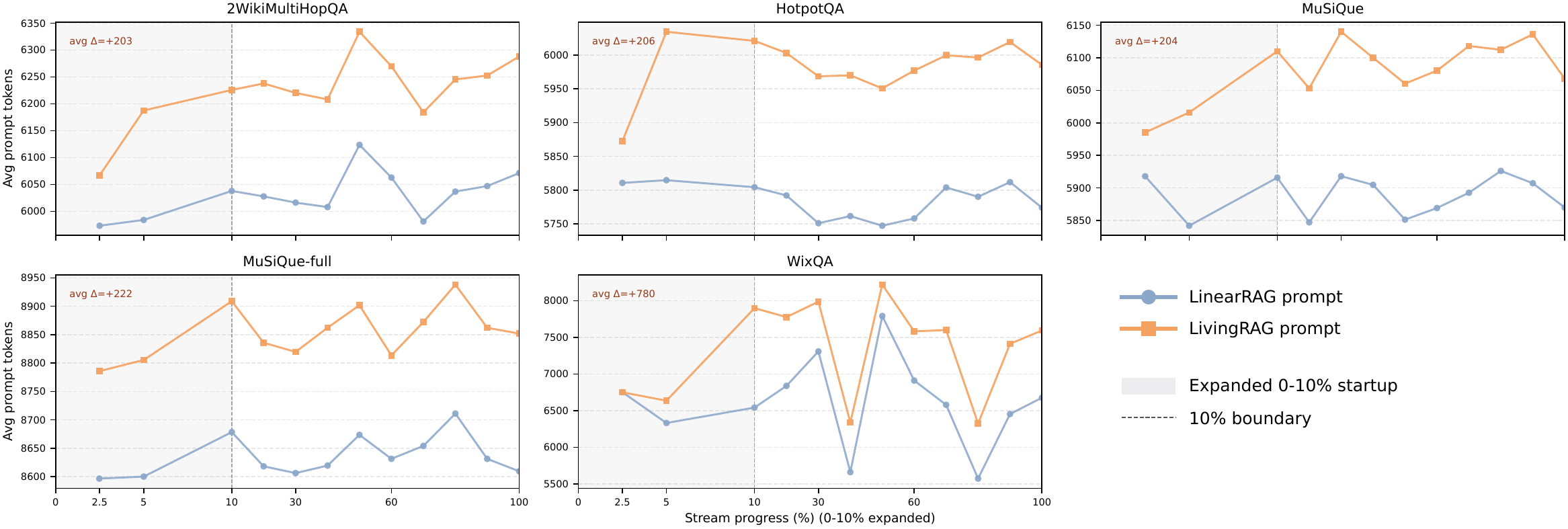}
\caption{
Prompt token trends over online QA streams.
Each panel corresponds to one dataset.
The blue curve shows LinearRAG and the orange curve shows LivingRAG.
The first 10\% of each stream is expanded to show the startup stage.
}
\label{fig:prompt_token_trend}
\end{figure*}

Figure~\ref{fig:prompt_token_trend} complements the completion token analysis in Figure~\ref{fig:completion_token_trend}.
It shows the input cost of LinearRAG and LivingRAG over the online stream.
At the very beginning, LivingRAG has not yet stored useful experiences.
It also has no reasoning scaffold to add to the prompt.
Thus, its prompt is the same as the LinearRAG prompt for the earliest queries.
After experiences are written, LivingRAG can select a reasoning scaffold and place it in the input context.
The selected scaffold makes the LivingRAG prompt longer than the LinearRAG prompt in later segments.
The trend is therefore consistent with Table~\ref{tab:token_efficiency}.
LivingRAG pays a small input token increase, but this increase is much smaller than the reduction in completion tokens.
Appendix~\ref{app:experience-use} further shows that scaffold use increases after the startup stage.

\section{Experience Use During Online Inference}
\label{app:experience-use}

\begin{figure*}[t]
\centering
\includegraphics[width=\textwidth]{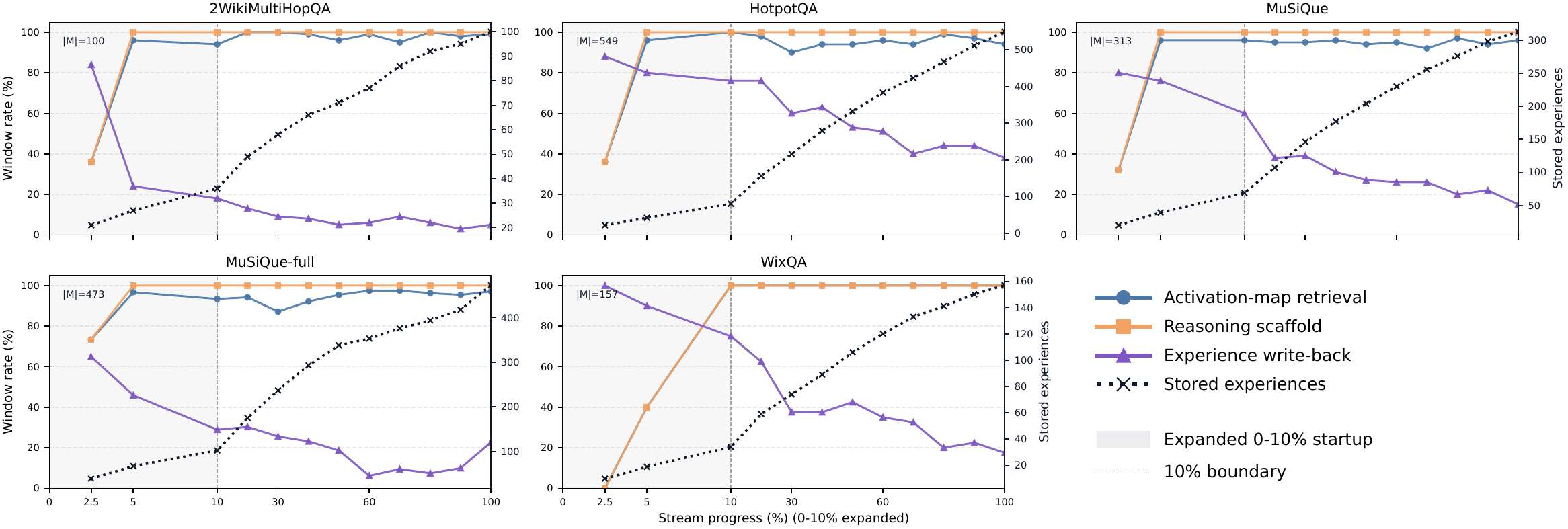}
\caption{
Online use of stored experiences in LivingRAG.
The curves show activation-map retrieval, reasoning scaffold use, experience write-back, and the cumulative number of stored experiences.
The first 10\% of each stream is expanded to show the startup stage.
}
\label{fig:experience_use_dynamics}
\end{figure*}

\begin{table*}[t]
\centering
\small
\setlength{\tabcolsep}{4pt}
\caption{Coverage of stored experiences during LivingRAG inference. Activation-map retrieval counts how often stored experiences are selected in the top-$K$ experience match set used for retrieval guidance. Reasoning scaffold reuse counts how often stored experiences are selected as generation scaffolds. Coverage is the percentage of stored experiences selected at least once.}
\label{tab:experience_source_coverage}
\begin{tabular}{lrrrrrr}
\toprule
\multirow{2}{*}{\textbf{Dataset}} & \multirow{2}{*}{\textbf{Stored experiences}} & \multicolumn{2}{c}{\textbf{Activation-map retrieval}} & \multicolumn{3}{c}{\textbf{Reasoning scaffold reuse}} \\
\cmidrule(lr){3-4} \cmidrule(lr){5-7}
& & \textbf{Top-$K$ selections} & \textbf{Coverage (\%)} & \textbf{Uses} & \textbf{Coverage (\%)} & \textbf{Max uses} \\
\midrule
2Wiki & 100 & 4820 & 97.00 & 984 & 85.00 & 103 \\
HotpotQA & 549 & 2814 & 79.60 & 984 & 58.11 & 20 \\
MuSiQue & 313 & 2799 & 86.90 & 983 & 70.29 & 30 \\
MuSiQue-full & 473 & 6828 & 93.02 & 2401 & 81.61 & 68 \\
WixQA & 157 & 1910 & 88.54 & 384 & 67.52 & 20 \\
\bottomrule
\end{tabular}
\end{table*}

Figure~\ref{fig:experience_use_dynamics} and Table~\ref{tab:experience_source_coverage} describe how LivingRAG uses experiences after they are written.
The figure shows that the use of stored experiences rises quickly after the startup stage.
Activation-map retrieval and reasoning scaffolds are used for most later queries in all datasets.
At the same time, experience write-back is more selective.
The pattern matches the design of LivingRAG.
The system can reuse earlier experiences often, but it only writes a new experience when the candidate is grounded and sufficiently novel.
The table further shows that reuse is spread across many stored experiences.
For activation-map retrieval, the coverage is above 79\% on every dataset.
For reasoning-scaffold reuse, the coverage is also broad, ranging from 58.11\% on HotpotQA to 85.00\% on 2Wiki.
These results indicate that the stored experiences are not only accumulated.
They are repeatedly selected by the retrieval and generation paths during later inference.
Appendix~\ref{app:direct-case-study} gives concrete examples of these reuse patterns.

\section{Direct Case Study: LinearRAG vs. LivingRAG}
\label{app:direct-case-study}

We next give aligned case studies from the final runs.
Each case compares the same stream index under LinearRAG and LivingRAG.
The first two cases show how activation-map fusion changes retrieval toward the answer-bearing evidence.
The last two cases show how a selected scaffold gives the LLM a reusable reasoning form and sharply shortens generation.

\begin{papertextbox}{E1: Experience-augmented retrieval recovers a missing Cabot chain}
\textbf{Dataset / index:} MuSiQue-full / 2231

\textbf{Question:} Whose navigator father explored the east coast of the region where Ignacio Esparza was later born?

\textbf{LinearRAG output:} Francisco de Orellana. \textbf{Result:} incorrect. \textbf{Completion tokens:} 8,739.

\textbf{LivingRAG output:} Sebastian Cabot. \textbf{Result:} correct. \textbf{Completion tokens:} 1,264.

\textbf{Experience retrieved by LivingRAG:}
Source question: Who is the child of the Italian navigator who sailed for England and explored the eastern coast of the continent Francisco Bojado's birthplace is located?
Stored answer: Sebastian Cabot.
Activation signal: 20 experience entities are added, and 16 new entities cross the retrieval threshold.
Important activated entities include Sebastian Cabot, John Cabot, Giovanni Caboto, Vinland, and continental North America.

\textbf{Retrieved articles and answer evidence:}
Ignacio Esparza: Esparza was born in Guadalajara, Jalisco, Mexico.
John Cabot: John Cabot was a Venetian navigator and explorer whose 1497 voyage under Henry VII explored the coast of North America.
Sebastian Cabot (explorer): Sebastian Cabot was the son of John Cabot.

\textbf{Summary of the reasoning chain:}
The current question asks for the child of a navigator. Ignacio Esparza gives Guadalajara, and the Guadalajara evidence links the location to North America. The retrieved Cabot evidence identifies the relevant navigator as John Cabot, not Francisco de Orellana. Since the Sebastian Cabot article states that Sebastian Cabot was John Cabot's son, the answer is Sebastian Cabot.
\end{papertextbox}

In E1, the useful stored experience is not only a previous answer.
It injects a coherent Cabot-centered entity neighborhood, which moves retrieval from a generic explorer branch to the evidence chain that contains the gold answer.

\begin{papertextbox}{E2: Experience transfers the Central Europe--406 event}
\textbf{Dataset / index:} MuSiQue-full / 798

\textbf{Question:} At the end of which year did tribes from the area where Linear Pottery culture was found invade the Roman Empire?

\textbf{LinearRAG output:} 113 BC. \textbf{Result:} incorrect. \textbf{Completion tokens:} 4,360.

\textbf{LivingRAG output:} 406. \textbf{Result:} correct. \textbf{Completion tokens:} 847.

\textbf{Experience retrieved by LivingRAG:}
Source question: At the end of which year did tribes from the place where the Celts are said to have migrated from invade the Roman empire?
Stored answer: 406.
Activation signal: 11 experience entities are added, and 10 new entities cross the retrieval threshold.
Important activated entities include Suebi, Vandals, Alani, Central Europe, and 31 December 406.

\textbf{Retrieved articles and answer evidence:}
Linear Pottery (LBK) context: the relevant Neolithic evidence places the cultural area in Central Europe.
Galicia (Spain): in the early fifth century, tribes of Central Europe, including the Suebi, Vandals, and Alani, crossed the Rhine on 31 December 406.

\textbf{Summary of the reasoning chain:}
The question first resolves the Linear Pottery area to Central Europe. The reused experience points to the same historical event pattern: Central European tribes entering Roman territory. The Galicia passage gives the date explicitly as 31 December 406. Therefore the requested year is 406.
\end{papertextbox}

E2 shows the same mechanism in a different domain.
The earlier experience supplies event-specific entities, so LivingRAG retrieves the passage that directly states the answer year instead of following a distracting ancient-date association.

\begin{papertextbox}{S1: Scaffold reuse for the Moon River--Ondine template}
\textbf{Dataset / index:} MuSiQue-full / 627

\textbf{Question:} For what did the singer of Moon River in Breakfast at Tiffany's win a Tony?

\textbf{LinearRAG output:} Best Actress in a Play for Ondine. \textbf{Result:} correct. \textbf{Completion tokens:} 3,628.

\textbf{LivingRAG output:} Ondine. \textbf{Result:} correct. \textbf{Completion tokens:} 325.

\textbf{Selected scaffold:}
Source question: For what did the singer of the original version of Moon River win a Tony award?
Stored answer: Ondine.
Scaffold score: 0.598. Template score: 0.877. Semantic score: 0.798.

\textbf{Retrieved articles and answer evidence:}
Moon River: the song was originally performed by Audrey Hepburn in Breakfast at Tiffany's.
List of awards and honours received by Audrey Hepburn: Hepburn won the 1954 Tony Award for Best Performance by a Leading Actress in a Play in Ondine.

\textbf{Scaffold shown to the LLM:}
Resolve the Moon River performer as Audrey Hepburn.
Then use the Audrey Hepburn awards passage.
The Tony Award evidence points to Ondine.
Answer with the award object, not the whole intermediate chain.

\textbf{LLM response after scaffold reuse:}
The current question uses the same template as the stored case. Moon River in Breakfast at Tiffany's identifies Audrey Hepburn. Her awards passage states that the Tony Award was for Ondine. The answer is Ondine.
\end{papertextbox}

S1 illustrates scaffold reuse without an accuracy change.
LivingRAG still uses the same evidence chain, but the scaffold makes the decomposition explicit before generation, reducing completion tokens by an order of magnitude.

\begin{papertextbox}{S2: Scaffold reuse for a repeated Clarksville--Nashville distance hop}
\textbf{Dataset / index:} MuSiQue / 948

\textbf{Question:} How many miles are there between Nashville and the place in Tennessee where Mark Day was born?

\textbf{LinearRAG output:} 45. \textbf{Result:} correct. \textbf{Completion tokens:} 7,054.

\textbf{LivingRAG output:} 45 miles. \textbf{Result:} correct. \textbf{Completion tokens:} 1,042.

\textbf{Selected scaffold:}
Source question: How far is Hod Lisenbee's place of death in TN from Nashville?
Stored answer: 45 miles.
Scaffold score: 0.475. Template score: 0.692. Semantic score: 0.732.

\textbf{Retrieved articles and answer evidence:}
Mark Day (racing driver): Mark Day was born in Clarksville, Tennessee.
Tennessee: Clarksville is about 45 miles northwest of Nashville.

\textbf{Scaffold shown to the LLM:}
Resolve the person-specific Tennessee location first.
Then reuse the Clarksville-to-Nashville distance passage.
If the current person also maps to Clarksville, answer 45 miles.

\textbf{LLM response after scaffold reuse:}
Mark Day's birthplace is Clarksville, Tennessee. The Tennessee passage gives Clarksville as about 45 miles northwest of Nashville. The answer is 45 miles.
\end{papertextbox}

S2 shows that the scaffold can be reused even when the surface entity changes.
The old case binds Hod Lisenbee to Clarksville, while the new case binds Mark Day to Clarksville; the second hop and the answer form remain the same.

\section{Additional Robustness and Transfer Analyses}
\label{app:additional-robustness}

\subsection{Benefits by Realized Reuse Signal}
\label{app:reuse-benefit}

\begin{table}[t]
\centering
\caption{Performance changes conditioned on realized graph and template reuse.}
\label{tab:reuse-conditioned-benefit}
\resizebox{\columnwidth}{!}{%
\begin{tabular}{lccrr}
\toprule
\textbf{Dataset} & \textbf{Graph} & \textbf{Template} & \makecell{\textbf{$\Delta$LLM-} \\ \textbf{Acc.}} & \makecell{\textbf{$\Delta$ Completion} \\ \textbf{tokens}} \\
\midrule
2Wiki & -- & -- & +0.59 & -12.8 \\
      & -- & $\checkmark$ & +2.97 & -189.1 \\
\midrule
MuSiQue-full & -- & -- & +0.56 & +23.6 \\
             & $\checkmark$ & -- & +2.12 & -297.2 \\
             & -- & $\checkmark$ & +4.73 & -1352.9 \\
             & $\checkmark$ & $\checkmark$ & +9.81 & -964.7 \\
\midrule
WixQA & -- & -- & +1.03 & -155.6 \\
      & $\checkmark$ & -- & +4.23 & -272.9 \\
      & $\checkmark$ & $\checkmark$ & +8.39 & -454.8 \\
\bottomrule
\end{tabular}%
}
\end{table}

We further conduct a query-level analysis of realized graph and template reuse.
Table~\ref{tab:reuse-conditioned-benefit} reports paired changes; $\Delta=$ LivingRAG $-$ LinearRAG, and negative completion values indicate savings.

Either signal yields greater benefits than neither, while their combination can further strengthen the gains.
Missing rows contain no samples.

\subsection{Query-Order Robustness}
\label{app:query-order}

\begin{table*}[t]
\centering
\caption{Robustness across different online query orders.}
\label{tab:query-order-robustness}
\resizebox{\textwidth}{!}{%
\begin{tabular}{llrrrrr}
\toprule
\textbf{Dataset} & \textbf{Setting} & \textbf{Cont.-Acc.} & \textbf{LLM-Acc.} & \textbf{Completion tokens} & \textbf{Graph trans.} & \textbf{Template reuse} \\
\midrule
2Wiki & Paper & 79.60 / 82.70 & 84.40 / 87.00 & 1067.6 / 906.4 & 0.00 & 84.15 \\
& 3 orders & 79.13$\pm$0.42 / 83.33$\pm$2.03 & 83.80$\pm$1.22 / 89.40$\pm$5.14 & 1038.7$\pm$25.4 / 806.4$\pm$92.7 & 0.00$\pm$0.00 & 81.54$\pm$2.69 \\
MuSiQue-full & Paper & 41.00 / 44.39 & 52.71 / 58.42 & 2223.8 / 1643.7 & 82.42 & 50.44 \\
& 3 orders & 41.36$\pm$1.12 / 44.21$\pm$0.90 & 53.10$\pm$0.44 / 58.09$\pm$0.88 & 2231.8$\pm$26.5 / 1744.8$\pm$94.4 & 81.84$\pm$0.55 & 50.15$\pm$0.64 \\
WixQA & Paper & -- & 66.00 / 70.25 & 1653.8 / 1378.1 & 91.50 & 7.03 \\
& 3 orders & -- & 66.25$\pm$1.39 / 70.25$\pm$1.00 & 1689.4$\pm$38.6 / 1403.6$\pm$22.2 & 91.92$\pm$0.38 & 7.29$\pm$0.30 \\
\bottomrule
\end{tabular}%
}
\end{table*}

We evaluate the same complete 2Wiki, MuSiQue-full, and WixQA streams under two additional random query permutations that change only arrival order.
In Table~\ref{tab:query-order-robustness}, ``3 orders'' reports the mean and sample standard deviation over the paper order and two permutations.

Accuracy and token cells report LinearRAG / LivingRAG.
Accuracies and reuse rates are percentages; tokens are per-query averages.
LivingRAG achieves higher accuracy than LinearRAG on every reported metric and uses fewer completion tokens in every tested order.
The realized reuse patterns also remain stable after reordering.
These consistent results provide evidence that LivingRAG's benefits are not only driven by one favorable stream arrangement.

\subsection{Transfer to an Entity-Passage PPR Retriever}
\label{app:ppr-transfer}

\begin{table}[t]
\centering
\caption{Retrieval-level transfer on WixQA.}
\label{tab:ppr-transfer}
\resizebox{\columnwidth}{!}{%
\begin{tabular}{lrrr}
\toprule
\textbf{Setting} & \textbf{Article R@5} & \textbf{Hit@5} & \textbf{MRR@5} \\
\midrule
Native PPR & 62.42 & 68.00 & 42.74 \\
+ Experience fusion & 63.42 & 69.00 & 44.82 \\
\bottomrule
\end{tabular}%
}
\end{table}

We chose LinearRAG because it is a strong, lightweight retriever with an explicit entity state before PPR.
This allows us to isolate experience reuse without changing the corpus graph or ranking procedure.
To test compatibility beyond LinearRAG, we run a preliminary retrieval-level test on a separate entity-passage PPR retriever.
As the test isolates interface compatibility rather than end-to-end integration, we report retrieval rather than generation metrics.

The adapter treats verified states as sparse priors over target graph nodes, an interface also exposed by HippoRAG2~\citep{hippoRAG2}.
For native reset $\mathbf{r}_q=[\mathbf{r}_q^E;\mathbf{r}_q^P]$, our adapter uses
\begin{equation}
\mathbf{r}'_q =
\bigl[(1-\lambda_q)\mathbf{r}_q^E + \lambda_q\tilde{\mathbf{h}}_q;
\mathbf{r}_q^P\bigr],
\end{equation}
where $\tilde{\mathbf{h}}_q$ mixes earlier verified states.
PPR and ranking remain unchanged.

We use all 400 WixQA queries and 7,132 corpus chunks.
The graph uses cached entity annotations and maps only earlier quality-gated states.

R@5 measures top-five gold-article coverage, Hit@5 any gold hit, and MRR@5 the first-hit rank.
The gains provide preliminary evidence of transfer through this interface.

\end{document}